\documentclass[runningheads]{llncs}

\usepackage{eccv}

\usepackage{eccvabbrv}

\usepackage{graphicx}
\usepackage{booktabs}

\usepackage[accsupp]{axessibility}  % Improves PDF readability for those with disabilities.

\usepackage{hyperref}

\usepackage{orcidlink}

\usepackage{hyperref}
\usepackage{url}

\usepackage{booktabs}       % professional-quality tables
\usepackage{amsfonts}       % blackboard math symbols
\usepackage{nicefrac}       % compact symbols for 1/2, etc.
\usepackage{microtype}      % microtypography
\usepackage{xcolor}         % colors

\usepackage[textsize=tiny]{todonotes}
\usepackage{multirow}
\usepackage{diagbox}
\usepackage{makecell}
\usepackage[para,online,flushleft]{threeparttable}

\usepackage{amsmath}
\usepackage{amssymb}
\usepackage{mathtools}

\usepackage{listings}

\usepackage{wrapfig,lipsum,booktabs}
\usepackage{algorithm,algpseudocode}

\usepackage[dvipsnames,svgnames]{xcolor}

\newcommand\fang[1]{\textcolor{black}{#1}}
\newcommand\eccvfang[1]{\textcolor{black}{#1}}

\newcommand\meng[1]{\textcolor{black}{#1}}

\newcommand\ICLRrebuttal[1]{\textcolor{black}{#1}}

\begin{document}

\title{Contrastive Conditional–Unconditional \\Alignment for Long-tailed Diffusion Model} 

\titlerunning{CCUA: Long-tailed Diffusion Models Training}

\author{Fang Chen\inst{1}%\orcidlink{0000-0003-4899-8474}
\and
Alex Villa\inst{1}%\orcidlink{0009-0000-3422-8626}
\and
Gongbo Liang\inst{2}%\orcidlink{0000-0002-6700-6664} 
\and \\
Li Fuxin\inst{3}%\orcidlink{0000-0002-6700-6664} 
\and
Xiaoyi Lu\inst{4}%\orcidlink{} 
\and
Meng Tang\inst{1}%\orcidlink{0000-0002-2984-1343}
}

\authorrunning{Fang et al.}

\institute{University of California Merced
%\email{{fchen20, avilla49, mtang4}@ucmerced.edu}\\ 
\and
Texas A\&M University-San Antonio
%\email{gliang@tamusa.edu} \\ 
\and
Oregon State University
%\email{fuxin.li@oregonstate.edu}\\ 
\and
University of Florida
%\email{xiaoyilu@ufl.edu}
}
\maketitle

\begin{abstract}
Training data for class-conditional image synthesis often exhibit a long-tailed distribution with limited amount of images for tail classes. Such an imbalance causes mode collapse and reduces the diversity of synthesized images for tail classes.
For class-conditional diffusion models trained with imbalanced data, we aim to improve the diversity and fidelity of tail class images without compromising the quality of head class images.
We propose contrastive conditional-unconditional alignment (CCUA), which comprises two synergistic loss functions.
Our first loss is an Alignment Loss (AL) that aligns class-conditional generation with unconditional generation at large timesteps. Alignment loss makes the denoising process insensitive to class conditions for the initial steps, which enriches tail classes through knowledge sharing from head classes. 
Secondly, we diversify unconditional generation via an Unsupervised Contrastive Loss (UCL) to increase the distance/dissimilarity among synthetic images.
%Such regularization is optionally coupled with a standard trick of batch resampling to further diversify tail-class images.
% Conditional-unconditional alignment has been shown to enhance the performance of long-tailed GAN. We are the first to adapt such alignment to diffusion models. 
We combine the two losses to implicitly diversify conditional generation.
%which is different from other contrastive learning variants.
%We successfully leverage contrastive learning and conditional-unconditional alignment for class-imbalanced diffusion models. 
Our framework is easy to implement as demonstrated on both U-Net based architecture and Diffusion Transformer. Our method outperforms vanilla denoising diffusion probabilistic models, score-based diffusion model, and alternative contrastive methods for class-imbalanced image generation across various datasets, in particular ImageNet-LT with 256$\times$256 resolution.
\end{abstract}

\section{Introduction}
\label{sec:intro}

\begin{figure}[t!bhp]
\centering
\begin{minipage}{.89\columnwidth}
\centering
    \subfloat[SiT~\cite{ma2024sit}]{
      \label{sit_redwine} \includegraphics[height=4.1cm]{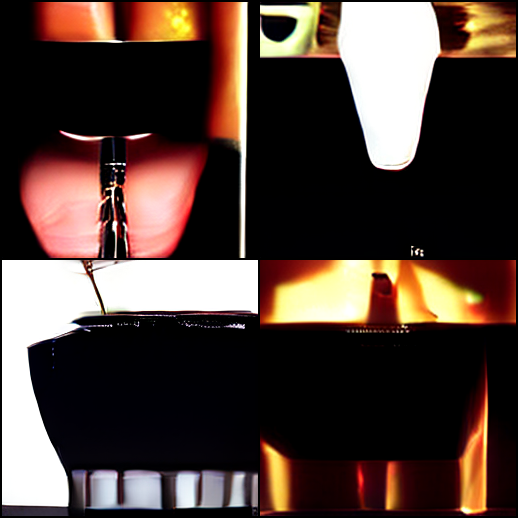}
      }
    \subfloat[SiT with CCUA (ours)]{
      \label{cldm_redwine} \includegraphics[height=4.1cm]{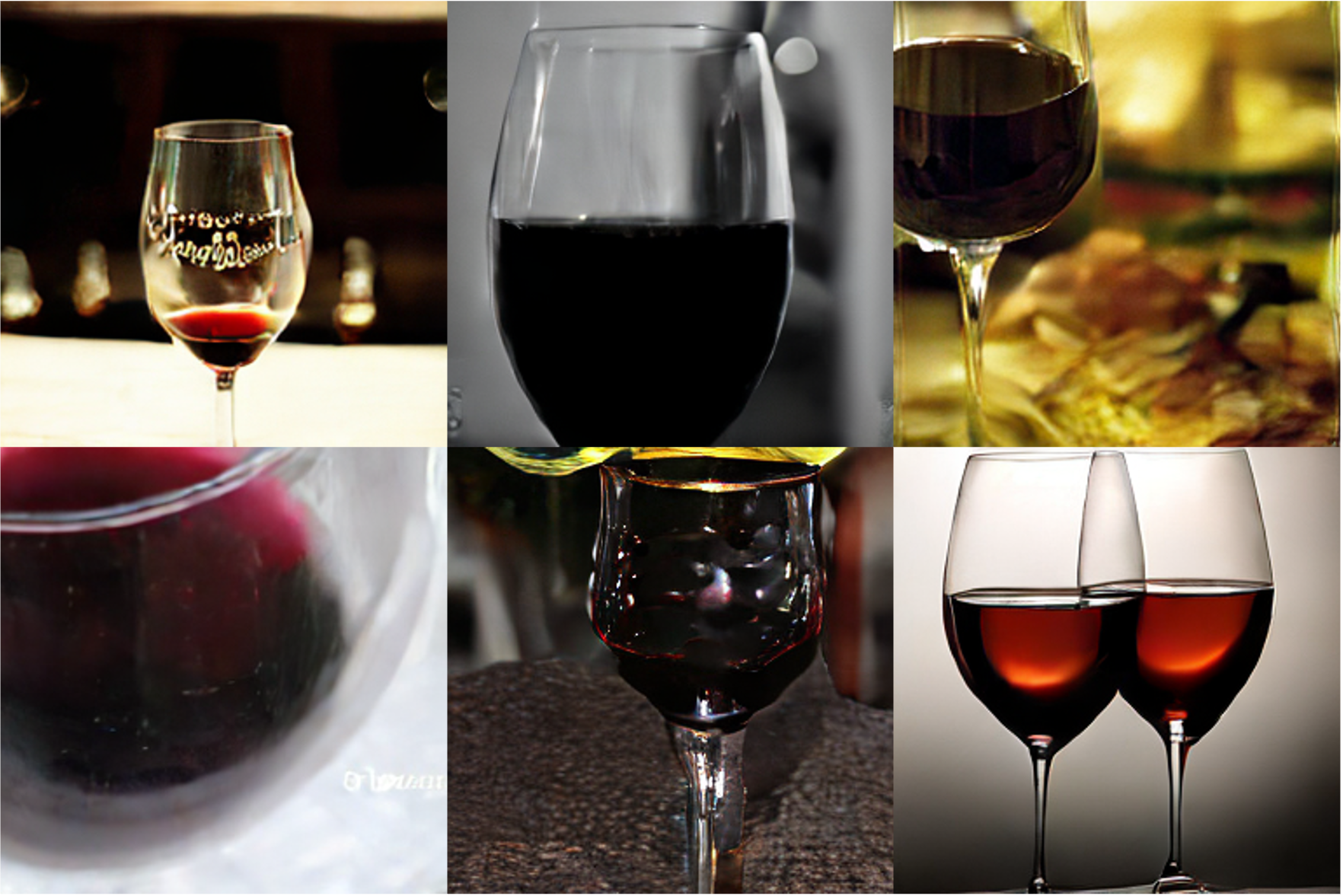}
      }
\end{minipage}

\caption{Generated images for a tail class (‘red wine’) with only 13 training images by (a) standard SiT~\cite{ma2024sit} with diffusion loss and (b) proposed CCUA framework. Both models are trained on long-tailed ImageNet dataset for 900k steps with 256x256 resolution.}
% %\vspace{-0.5cm}
\label{fig:intro}
\end{figure}

% Background
Recent advances in diffusion models~\cite{ddpm, ddim} have led to breakthroughs in various generation tasks such as image generation~\cite{zhang2023adding, stable_diffusion,ruiz2023dreambooth}, video generation~\cite{ho2022imagen, ho2022video}, image editing~\cite{couairon2022diffedit, kawar2023imagic}, 3D generation~\cite{poole2022dreamfusion}, etc.
%
%These models have not only established new state-of-the-art benchmarks but also enabled researchers to exert finer control over both the generation process and the quality of synthesized outputs.
%The improved architectures and training methodologies have facilitated more precise manipulation of latent representations, resulting in enhanced fidelity and diversity in generated samples across multiple domains.
%
These diffusion-based generative models rely on large-scale datasets for training, which often follow a long-tailed distribution with a significant amount of data for head classes and a limited amount of data for tail classes.
Similar to the class-imbalanced recognition model~\cite{liu2022open} and the class-imbalanced generative adversarial network~\cite{tan2020fairgen,rangwani2022improving,rangwani2023noisytwins,utlo}, diffusion model is often not able to generate high-quality images for tail classes due to the scarcity of training data.
%diffusion model also tends to \textit{overfit} on tail classes leading to mode collapse.
%
As shown in Fig.~\ref{fig:intro}, the original diffusion model with transformer backbone ~\cite{ma2024sit} generates inferior images for a tail class (red wine) when trained using a long-tailed version of ImageNet.
We aim to increase the fidelity and diversity of tail class images while maintaining the quality of head class images.% for diffusion models.
% %
% Our work has a positive societal impact on content generation of underrepresented groups.
% %

 % \begin{figure}[t]
 %     \centering
 %     \includegraphics[width=1.0\linewidth]{figures/intro.pdf}
 %     \caption{Given a long-tailed dataset with only five images for the tail class (worm), diffusion model }
 %     %\caption{The synthesized dog images of diffusion model trained by classifier-free guidance (CFG)~\cite{cfg} on an imbalanced dataset of 1.3k tabby cat images and 13 maltese dog images from ImageNet~\cite{imagenet}. a) Mode Collapse in Tail Classes: Due to insufficient dog training images, the model generates overly similar dog images. b) Head-Class Bias: The model produces samples oriented towards head-class `cat', i.e. `cat-like' dogs, even though under the `dog' class prompt guidance.}
 %     \label{fig:intro}
 % \end{figure}

%

\begin{figure}[thbp]
\centering
\includegraphics[width=0.62\columnwidth]{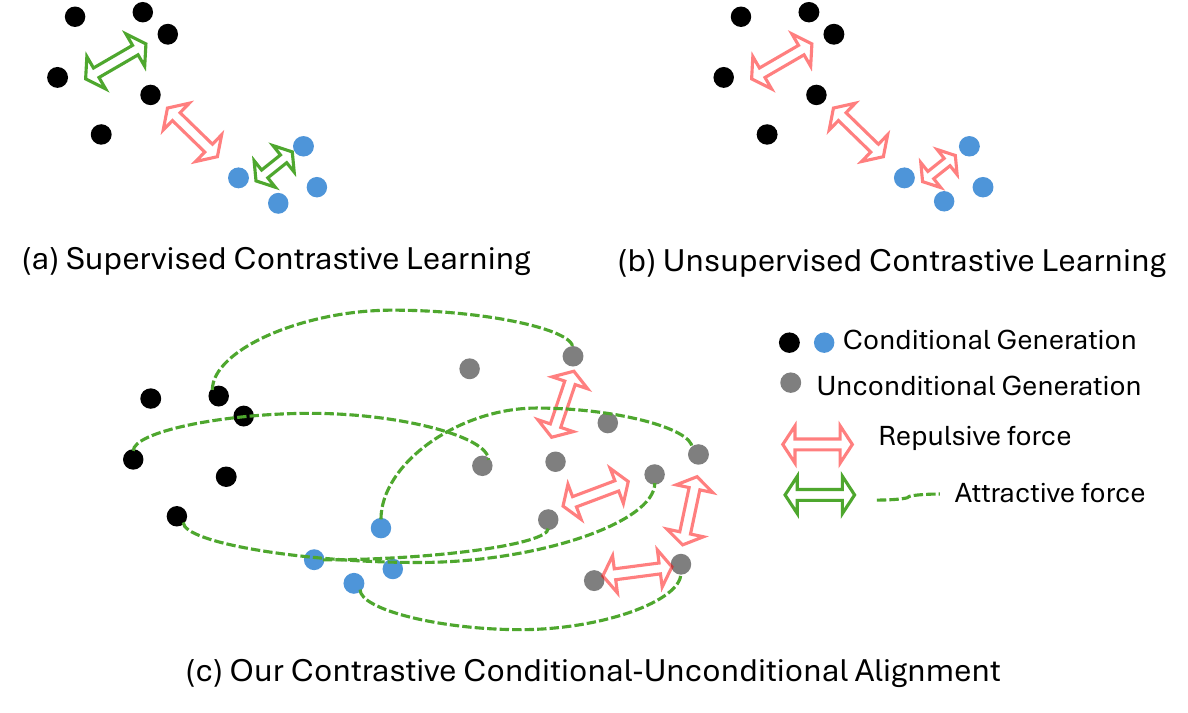}
\caption{Variants of contrastive learning for regularizing latents of conditional \& unconditional generation and our proposed CCUA framework. The black and blue dots are latents from class-conditional generation and grey dots are from unconditional generation. The latents represent intermediate features from a U-Net or transformer-based denoising network. Rather than applying standard supervised contrastive learning (a) and unsupervised contrastive learning (b) for \textit{conditional} generation, we propose to diversity \textit{unconditional} generation via repulsive force and further align \textit{conditional} generation with \textit{unconditional} generation via attractive force (c).
 }
\label{fig:ccua_concept}
\end{figure}

We look into a highly related and crucial problem of long-tailed image recognition, for which contrastive learning is an effective method~\cite{jiang2021self,li2022targeted}.
Rather than applying standard supervised contrastive learning ~\cite{khosla2020supervised} or unsupervised contrastive learning~\cite{infonce} for conditional generation, we propose
to diversify unconditional generation via a repulsive force among samples and further align the conditional
generation with unconditional generation via an attractive force.
Fig.~\ref{fig:ccua_concept} illustrates how our contrastive conditional-unconditional alignment framework is different from other contrastive learning variants.

Specifically, our framework combines two synergistic losses. Our first loss is an unsupervised contrastive loss with negative samples only to diversify unconditional generation, serving as regularization for denoising diffusion probabilistic models (DDPM) and score-based diffusion models (SBDM).
%
%Mode collapse during inference implies that generated samples are too close to each other,  particularly for tail classes.
Given that mode collapse during inference manifests as generated samples being overly similar for tail classes, we introduce unsupervised contrastive learning with negative samples to maximize the distances among generated images.
%
%Our loss defined on the bottleneck layer of UNet-based diffusion models is simple to implement. % and adds little computational overhead during training.
%
Unlike supervised contrastive learning~\cite{khosla2020supervised}, unsupervised contrastive learning distinguishes between representations of images regardless of their class, hence increasing intra-class and inter-class image diversity.
Our unsupervised contrastive loss is implemented with batch resampling, which is a standard strategy for long-tailed recognition and generation~\cite{oclt, zhang2019balance, shi2023re}.
%Batch resampling is a standard strategy for long-tailed recognition and generation~\cite{oclt, zhang2019balance, shi2023re}. However, oversampling tail classes generally makes the network prone to over-fitting, since head class samples are often consequently under-represented~\cite{zhang2019balance}.
%Our unsupervised contrastive loss relies on pairs of images, that batch resampling could increase the chance of images from the same tail class appearing in the same batch, which further diversify tail class images. We choose batch resampling as an optional strategy for our framework.

%
Our second loss is an alignment loss designed to align estimated noises from conditional and unconditional generation, which effectively minimizes the KL divergence between latent distributions for conditional and unconditional generation.
While such conditional-unconditional alignment seems undesirable with less controllability, a critical aspect is that our alignment loss is weighted more for larger timesteps corresponding to the initial stage of the reverse process.

Our alignment loss can also be motivated by the success of conditional-unconditional alignment for long-tailed GAN~\cite{utlo,transitioncgan}, which facilitates knowledge sharing between head classes and tail classes.
Notably, unconditional GAN generation has been observed to achieve superior FID than conditional generation under limited data~\cite{transitioncgan}.
%
%We adapt conditional-unconditional alignment from GAN to diffusion models.
Unlike the GAN-based method~\cite{utlo} which aligns conditional generation and unconditional generation for low-resolution representations of images exhibiting intra-class similarity, we propose to match conditional generation with unconditional generation for large timesteps, by leveraging observed image similarity during the initial denoising steps for tail classes and head classes~\cite{si2024freeu}.

The synergy of the two losses makes our method effective, with the unsupervised contrastive loss diversifying \textit{unconditional} latents repulsing negative pairs, and alignment loss aligning \textit{conditional} and \textit{unconditional} latents. As a result, conditional latents are diversified implicitly, which is theoretically and empirically better than directly diversifying conditional latents.
Previous work utilized contrastive learning to improve adversarial robustness~\cite{ouyang2023improving}, find semantically meaningful directions~\cite{dalva2024noiseclr}, accelerate training~\cite{repa}, and regularize representation~\cite{wang2025diffuse}. We effectively leveraged alignment and contrastive learning for class-imbalanced diffusion models and demonstrated the superior performance of our method on long-tailed image generation via comprehensive experiments.
% Most relevant is concurrent work of Dispervise Loss~\cite{wang2025diffuse} that integrates contrastive loss as regularization.
% The dispersive loss is directly and explicitly applied to conditional training, as designed for balanced diffusion models.
% In contrast, our InfoNCE loss is formulated in the unconditional latent space and is implicitly extended to conditional training through our second loss for conditional–unconditional alignment.
% Experiments demonstrated the superior performance of our method over Dispersive Loss~\cite{wang2025diffuse} on training long-tailed diffusion model.

% %
% Most relevant is concurrent work of Dispervise Loss~\cite{wang2025diffuse} that integrates contrastive loss as regularization. 
% While our first loss with InfoNCE is very similar to Dispersive Loss, we also propose a second loss for conditional-unconditional alignment. 
% Experiments demonstrated the superior performance of our method over Dispersive Loss~\cite{wang2025diffuse}.
%
%What's more, we study contrastive learning as a loss for training which is different from the contrastive guidance for inference~\cite{ouyang2023improving}.

The main contributions of this work are as follows:

\begin{itemize}
  \item We propose \textbf{C}ontrastive \textbf{C}onditional-\textbf{U}nconditional \textbf{A}lignment (\textbf{CCUA}) for Diffusion Model with imbalanced data. Our proposed losses are easy to implement with DDPM and SBDM pipelines for both UNet-based architecture and Diffusion Transformer.
  \item Firstly, our Alignment Loss (AL) aligns unconditional generation and conditional generation for the initial steps in the denoising process, facilitating knowledge sharing between head and tail classes.
  \item Secondly, our Unsupervised Contrastive Loss (UCL) employs unsupervised contrastive learning with negative samples only, enhancing intra-class diversity for unconditional generation.
  \item We improved the diversity and fidelity of tail class for conditional generation while maintaining the quality of head class for multiple datasets and various resolutions, in particular ImageNet-LT with 256x256 resolution. Our framework outperforms other contrastive learning variants, such as the concurrent work of dispersive loss~\cite{wang2025diffuse}.
  %We also improve results for unconditional generation.
  %\item Our method outperforms existing methods for long-tailed diffusion models as demonstrated across multiple datasets with various image resolutions including CIFAR10-LT, CIFAR100-LT, and ImageNet-LT.
\end{itemize}
%

%

%

% The development of high-fidelity diffusion models for image synthesis necessitates extensive training datasets, which frequently exhibit long-tailed distributions when sourced from real-world scenarios~\cite{cbdm, diffrop}.
%This inherent data imbalance presents significant challenges in model training, as it can lead to suboptimal performance in underrepresented classes or rare instances. 
%Consequently, addressing the implications of long-tailed distributions in training data has become a crucial consideration in advancing the robustness and generalization capabilities of diffusion-based image synthesis models.

%In long-tailed datasets, the primary challenge for training diffusion models lies in the scarcity of tail class samples, which impedes the achievement of uniform generative quality across head and tail classes~\cite{cbdm, diffrop}.
%This data imbalance typically results in limited representation learning for tail classes, leading to two significant issues in the generated outputs as shown in Fig.~\ref{fig:intro}: 
%a) Mode collapse in tail classes: Due to insufficient training data for tail classes, the model may exhibit a tendency to generate repetitive or overly similar samples for tail classes, failing to capture the full diversity within these underrepresented categories.
%b) Head-class bias: The model tends to produce samples that are disproportionately oriented towards head classes, potentially misrepresenting or overlooking the nuances of tail classes.

\section{Related Work}
\label{sec:related work}
\textbf{Class-imbalanced Image Generation}\quad 
Generative models such as VAE~\cite{vae}, GAN~\cite{stylegan,biggan}, and diffusion models~\cite{ddpm,stable_diffusion} generate inferior images for tail classes when trained with real-world data with a long-tailed distribution.
It has attracted a lot of research interest~\cite{ai2023generative,tan2020fairgen,utlo,cbdm,oclt} to address this issue for various types of models.
Many methods address the problem of class imbalance by augmenting training data for the tail classes.
A VAE is fine-tuned on tail classes under a majority-based prior~\cite{ai2023generative}.
%
% It is observed that GAN~\cite{tan2020fairgen} can amplify biases leading to tail class images barely generated during inference, and the fairness of GAN needs improvement.
It is observed that GAN~\cite{tan2020fairgen} can amplify biases, leading to tail classes to be barely generated during inference, highlighting that the fairness in GAN needs improvement.
Khorram et al.~\cite{utlo} propose a GAN-based long-tailed generation method, named UTLO, which shares the latent representations of conditional GAN with unconditional GAN and implicitly shares knowledge between the head class and tail class.
The motivation with UTLO is that low-resolution representations of images from GAN are similar for head classes and tail classes.
We observe a similar phenomenon for denoised images in the initial steps of the denoising process, and further propose a conditional-unconditional alignment loss designed for diffusion models.
Recent work addresses class-imbalanced diffusion models~\cite{cbdm,diffrop,oclt} by regularization losses to align or separate the distributions of synthetic images and their corresponding latent representations across different classes.
For example, CBDM~\cite{cbdm} loss minimizes the distance of estimated noise items between these two models, the original DDPM model and a second model trained with pseudo labels which form a uniform distribution.
DiffROP~\cite{diffrop} attempts to combine contrastive learning with diffusion model by maximizing the distance of distributions between classes.
% %
% %However, DiffROP barely improves the results over standard DDPM.
% %
% We aim to reduce overfitting for tail classes, while DiffROP~\cite{diffrop} reduces overlapping between different classes.
% %
% %We found the more pressing problem is intra-class overfitting rather than inter-class overlapping.
% %
% The first part of our loss, the InfoNCE loss with negative pairs only, is defined for any pair of images regardless of their classes.
% %
However, DiffROP only considers pairs of images of different classes as negative pairs without regularizing images of the same class.

\noindent \textbf{Contrastive Learning for Representation Learning}\quad
Both unsupervised contrastive learning~\cite{chen2020simple,moco} and supervised contrastive learning~\cite{khosla2020supervised,jiang2021self,li2022targeted} are representation learning methods by maximizing the similarity between positive pairs and minimizing the similarity between negative pairs.
Unsupervised contrastive learning does not require labels and typically augments the same data to form positive pairs~\cite{chen2020simple,moco}.
Supervised contrastive learning incorporates class labels and pulls together all samples from the same class while pushing apart samples from different classes.
%
%Contrastive learning is an effective method for self-supervised learning~\cite{moco} and long-tailed recognition~\cite{jiang2021self,li2022targeted}.
%
We propose a new variant of contrastive learning with conditional-unconditional alignment for class-imbalanced diffusion models.
Our work is different from the contrastive-guided diffusion process~\cite{ouyang2023improving}, which aims to improve adversarial robustness.
%
%Another difference is that we focus on contrastive learning during training, while contrastive-guided diffusion process~\cite{ouyang2023improving} is about guidance for inference.
%
While the diffusion model originally for generation tasks becomes an emerging technique for representation learning~\cite{fuest2024diffusion,baranchuk2021label}, we directly integrate contrastive learning regularization to diffusion models for better generation on imbalanced data.
%
%Contrastive learning is a representation learning method where the goal is to learn meaningful feature representations by contrasting positive and negative sample pairs~\cite{le2020contrastive, hu2024comprehensive}.
%
%In deep learning, contrastive learning is self-supervised representation learning by training a model to differentiate between similar and dissimilar samples~\cite{le2020contrastive, hu2024comprehensive}.
%
%Common contrastive learning losses include triplet loss, N-pair loss, and InfoNCE loss. 
Notably, unsupervised contrastive loss with negative samples exclusively focuses on separating dissimilar instances in the embedding space, which we leverage to address mode collapse, as detailed in Sec.~\ref{sec:ucl}.
% %
% BYOL~\cite{BYOL} is a contrastive learning method with MSE loss for positive pairs only similar to our loss formulation in Sec.~\ref{sec:al}.
%
In the context of generative modeling, REPA~\cite{repa} aligns diffusion model features with features from a frozen vision encoder to accelerate training. Unlike REPA, we directly regularize diffusion model features without external models.
Most relevant is concurrent work of dispersive loss~\cite{wang2025diffuse}, which is similar to our unsupervised contrastive loss with negative pairs only.
Dispersive loss is directly and explicitly applied to conditional training, as designed for balanced diffusion models.
By comparison, our unsupervised contrastive loss is formulated in the unconditional latent space and is implicitly extended to conditional training through our conditional–unconditional alignment loss.
This carefully designed contrastive conditional-unconditional alignment framework for long-tailed diffusion models achieves better generation as shown in experiments.
% However, our contrastive conditional-unconditional alignment framework for long-tailed diffusion models further aligns conditional and unconditional generation, which achieves better generation as shown in experiments.

\section{Method}
\label{sec:method}

We propose \textbf{C}ontrastive \textbf{C}onditional-\textbf{U}nconditional \textbf{A}lignment (CCUA) for Diffusion Model.
We show the intuition and motivation in Sec.~\ref{sec:ccua_motivation}.
Our method involves a unsupervised contrastive loss for unconditional generation and a conditional-unconditional alignment loss, as detailed in Sec.~\ref{sec:ucl} and~\ref{sec:al}.
Sec.~\ref{sec:framework} summarizes our framework and discusses the synergy between the two losses.

\subsection{Motivation of CCUA Framework}
\label{sec:ccua_motivation}

% main issue
Limited amount of training images for tail classes leads to limited diversity for synthetic images.
% sampling with replacement
Batch sampling with replacement leads to more severe overfitting on training tail class samples.
We explore contrastive learning, which is shown effective for long-tailed recognition~\cite{khosla2020supervised,jiang2021self,li2022targeted}, for the new task of long-tailed generation.
The idea is to regularize latents of a denoising network at timestep $t$, which can be intermediate feature or output noise from a U-Net or a transformer.
We discuss variants of contrastive learning as regularization for latents, and how our proposed framework is different.

% supervised contrastive learning
Supervised contrastive learning~\cite{khosla2020supervised} as shown in Fig.~\ref{fig:ccua_concept} (a) employs an attractive force for latents of the same class and a repulsive force for latents of different classes.
%
% unsupervised contrastive learning, replusive force will diversity
% We propose a synthergetic method
% we discuss the limitatin in 
\begin{figure}[t!bhp]
\centering
\includegraphics[width=0.6\columnwidth]{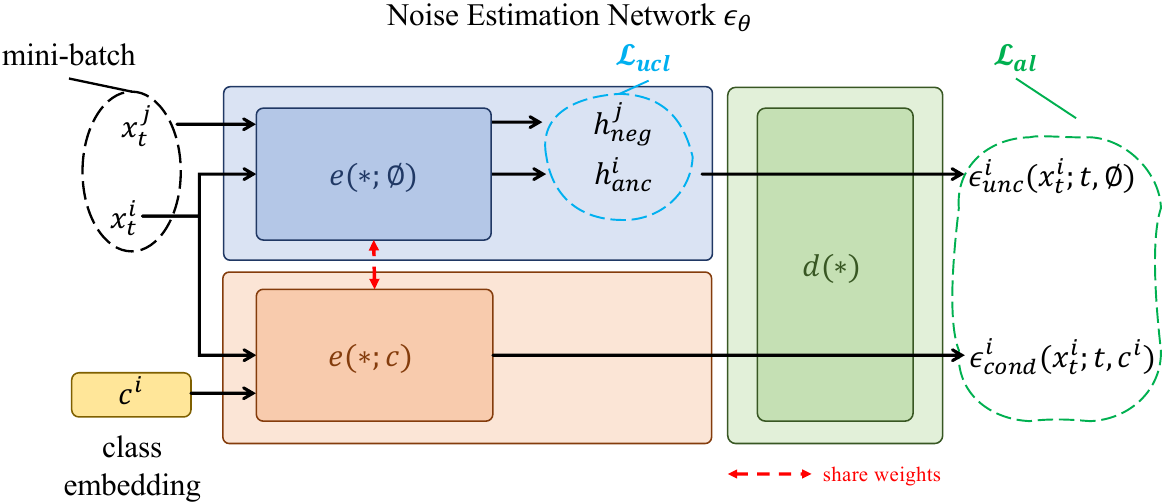}
\caption{Implementation of the proposed CCUA framework.
The noise estimation network is divided into a latent encode network $e(*)$ and a decode network $d(*)$.
$e(*)$ encodes input $x_t$ to a low-dimensional latent $h$, which is decoded to noise $\epsilon$ by $d(*)$.
 % %
 % The encoder of the noise estimation network $e(\cdot)$ encodes an image with noise $x_t$ to a low-dimensional latent $h$, which is decoded to noise $\epsilon$ by the decoder $d(\cdot)$.
 % %
 We increase the distance of unconditional latents by $\mathcal{L}_{ucl}$ with negative samples only, and align unconditional and conditional generation from the same sample $x_t^i$ and utilize $\mathcal{L}_{al}$ to minimize their distances at initial time steps. Specifically, $e(*)$ and $d(*)$ for unet-based model and diffusion transformer are shown in Fig.~\ref{fig:method_details} in Appendix~\ref{sec:details}.%\TODO{make b and c smaller} \TODO{discuss b and c, how intermediate embedding was selected}
 }
\label{fig:method}
\end{figure}
%A more reasonable alternative for generative models is unsupervised contrastive learning with repulsive force for any pair of images within a batch ( Fig.~\ref{fig:ccua_concept} (b)), which can be implemented via InfoNCE loss~\cite{infonce} with negative samples only, as is done in the concurrent work of dispersive loss~\cite{wang2025diffuse}.
%
However, supervised contrastive learning doesn't improve \textbf{intra-class diversity} as it involves an attractive force for pairs of images of the same class.
A better way to improve diversity is to have repulsive force for all intra-class images and inter-class images, which can be implemented via InfoNCE loss~\cite{infonce} with negative samples only in the concurrent work of dispersive loss~\cite{wang2025diffuse}.

We reveal the theoretical limitations of such unsupervised contrastive learning on conditional generation in Sec.~\ref{sec:ucl}.
We propose a dramatically different method that diversifies unconditional latents and also aligns conditional latents with unconditional latents shown in Fig.~\ref{fig:ccua_concept} (c). A repulsive force on unconditional latents improves diversity in particular for tail classes, and an attractive force aligns conditional generation with unconditional generation for large timesteps.
The two losses combined implicitly diversify latents from conditional generation.
Fig.~\ref{fig:method} gives an overview of how our framework is implemented.
%while Appendix~\ref{sec:details} details how we apply CCUA framework to UNet-based models and Diffusion Transformers.

%\TODO{mention applicability to both UNet and Transformer}
%
% As is common in contrastive learning, our framework involes both negative and positive pairs, as outlined in Sec.~\ref{sec:ucl} and Sec.~\ref{sec:al}, respectively. Sec.~\ref{sec:framework} summarizes our overall framework.

%Our contrastive loss consists of two parts, the infoNCE loss which is defined in the latent space, and the MSE loss which is defined in the $\epsilon$-space. 

\subsection{Unsupervised Contrastive Loss for Unconditional Generation}
\label{sec:ucl}

We observed the synthesized images of DDPM for a tail class concentrated around the limited training images in the latent space, leading to mode collapse, as shown in Fig.~\ref{fig:distribution}, and Fig.~\ref{fig:nearest-neighbor} in Appendix~\ref{sec:mode collapse issue}.

We consider contrastive loss with negative samples only, where the negative samples are based on noisy images from a mini-batch.
We treat a synthetic image and itself as a positive pair and all other images within the batch as negative samples.
By applying the contrastive loss with negative samples only, we increase the distance of each synthesized image from other images in the latent space, hence increasing the diversity of synthetic images in particular for tail classes.
Specifically, we divide a noise estimation network into two parts, a latent encode network $e(*)$, and decode network $d(*)$.
$e(*)$ encodes an image with noise $x_t$ to a low-dimensional latent $h$, which is decoded to the estimate noise item $\epsilon$ by $d(*)$.
Our contrastive loss is defined for the latents $h$ with each image in a mini-batch treated as an anchor.
All other images are treated as negative samples.
The anchor sample $x_t^i$ and negative samples $x_t^j$ are fed into the encoder $e(*)$ to get unconditional latents $h_{anc}^i = e_{\theta}(x_t^i;t, \emptyset)$ and $h_{neg}^j=e_{\theta}(x_t^j;t, \emptyset)$.
Our unsupervised contrastive loss $\mathcal{L}_{ucl}$ is defined as follows:

 \begin{equation}\label{eq:nce loss}
 \begin{aligned}
 % \mathcal{L}_{ucl}=-\frac{1}{|B|}\sum_{i\in B}{\log\frac{exp(h_{anc}^i \cdot h_{pos}^i/\tau)}{\sum_{j\in{B}}{exp(h_{anc}^i \cdot h_{neg}^j/\tau)}}},
 \mathcal{L}_{ucl}=&-\frac{1}{|B|}\sum_{i\in B}{\log\frac{\pi^i_{anc}}{\pi^i_{anc} + \sum_{j\in{B}, j \neq i}{\pi^j_{neg}}}}, \\
 &\text{with } \pi^i_{anc} = \exp(\frac{h_{anc}^i \cdot h_{pos}^i}{\tau})= \exp(\frac{1}{\tau}),
 \pi^j_{neg} = \exp(\frac{h_{anc}^i \cdot h_{neg}^j}{\tau}),
 \end{aligned}
 \end{equation}where $B$ denotes a mini-batch, and $\tau$ is a temperature for softmax which we keep $0.1$ as the default setting.
 Note that we do not augment an anchor image $x_t^i$ to be a positive sample like many unsupervised contrastive learning methods. Instead, we directly treat the anchor latent $h_{anc}^i$ itself as the positive vector in the contrastive loss, i.e., $h_{pos}^i:=h_{anc}^i$.
 We also normalized latents following standard protocol for InfoNCE loss~\cite{infonce}.
 We discuss implementation details on batch sampling for our UCL loss in Appendix~\ref{sec:details}.

\noindent \textbf{Why unsupervised contrastive loss improves image diversity?} In the worst case, that all embeddings collapse to a constant vector, the contrastive loss becomes $\log |B|$ for a batch with $|B|$ samples.
Such constant embeddings yield the maximum possible loss and are therefore discouraged. 
Intuitively, contrastive loss introduces a repulsive force between different samples.
The numerator remains constant, as self-similarity is always maximal, while the denominator aggregates pairwise similarities across the batch. Minimizing the loss requires reducing similarities between distinct samples, effectively pushing their embeddings apart in feature space. 
%
%In this way, the model avoids collapse by maximizing inter-sample distances. 
Geometrically, the optimal solution corresponds to embeddings being uniformly distributed on a hypersphere.

% \TODO{Add a paragraph on why InfoNCE loss with negative pairs only enhances diversity but avoid trivial solutions: Suppose all embeddings collapse to a constant vector. For a batch with N samples, the InfoNCE loss becomes, which is actually the worst possible case. Thus, constant embeddings give the highest possible loss and are avoided. Intuitively, InfoNCE creates a repulsive force between different samples: The numerator ⁡is constant since self-similarity is maximal. The denominator contains all pairwise similarities. Minimizing the loss requires reducing similarities between different samples, which forces embeddings of different samples to spread apart in feature space. The model avoids collapse by maximizing inter-sample distances. There is also a geometric interpretation to it. The optimal solution (without augmentation) spreads embeddings uniformly on a hypersphere.}

\noindent \textbf{Limitation of diversifying conditional latents directly} \ICLRrebuttal{It is known that InfoNCE loss~\cite{infonce} for contrastive learning gives a lower bound of mutual information between data sample and representation/latents. In other words, minimizing InfoNCE loss maximizes mutual information. We denote conditional latent as $h^{c}_{t} := e_\theta(x_t,t, c)$ and unconditional latent as $h^{u}_{t} := {e}_\theta(x_t,t,\emptyset)$, where $c$ is the class condition.
 The InfoNCE loss on unconditional latents maximizes the mutual information $I(h_{t}^{u};x)$ between image $x$ and its unconditional latent $h_t^u$. Similarly, InfoNCE loss on conditional latents ~\cite{wang2025diffuse} maximizes the mutual information $I(h_{t}^c;x,c)$ between image $x$ and its conditional latent $h_t^c$ and condition $c$.}

Based on the chain rule of mutual information, $I(h_{t}^c;x,c)$ can be decomposed into two parts:

\begin{equation}
\begin{aligned}
I(h^{c}_{t};x,c)=I(h^{c}_t;c)+I(h^{c}_t;x|c).
\end{aligned}
\end{equation}

\ICLRrebuttal{The first item $I(h^c_t;c)$ measures the mutual information between class $c$ and latent $h_t^c$, while the second item $I(h^{c}_{t};x|c)$ measures the conditional mutual information.
Intuitively, the first term tells how much class condition reveals about the latent, which reflects \textbf{inter-class diversity}. The second term measures conditional mutual information with condition $c$ which reflects \textbf{intra-class diversity}. 
To address the mode collapse issue for tail classes and increase diversity, it is apparent that we need to focus on maximizing the second item, i.e., conditional mutual information $I(h^{c}_{t}; x|c)$.
However, the optimization of inter-class mutual information $I(h_t^c;c)$ admits a trivial solution of having an identical latent for all images of the same class leading to maximum mutual information.}

\noindent \textbf{Unconditional Contrastive Loss and Combination with Alignment Loss}\quad We choose to diversify unconditional latents through InfoNCE loss with negative samples only, which maximizes $I(h^{}_t;x)$. In this case, there is no shortcut solution, and the model is forced to diversify all latents regardless of class. We further distill diversified unconditional latent to conditional latent with a simple alignment loss discussed in Section~\ref{sec:al}.

\subsection{Conditional-unconditional Alignment Loss}
\label{sec:al}

As shown in Fig.~\ref{fig:ccua_concept}, one of our goals is to align the distribution of latents from conditional and unconditional generation, which implicitly diversify conditional generation. Specifically, we penalize KL divergence between conditional distribution $p_\theta(x_{t-1}|x_t^i,c^i)$ and unconditional distribution $p_\theta(x_{t-1}|x_t^i)$:
\begin{equation}\label{eq:KL divergence of mse loss}
\begin{aligned}
\mathcal{L}^{i,t}_{al}=D_{KL}[p_\theta(x_{t-1}|x_t^i,c^i)||p_\theta(x_{t-1}|x_t^i)].
\end{aligned}
\end{equation}Suppose
\begin{equation}
\begin{aligned}
 p_\theta(x_{t-1}|x_t,c)&=\mathcal{N}(x_{t-1};\mu_\theta(x_t,t,c),\sigma_t^2I), \\
 p_\theta(x_{t-1}|x_t)&=\mathcal{N}(x_{t-1};\mu_\theta(x_t,t),\sigma_t^2I),
\end{aligned}
\end{equation}then we have:
\begin{equation}\label{eq:expectation form of mse loss}
\begin{aligned}
\mathcal{L}^{i,t}_{al} = \mathbb{E}[\frac{1}{2\sigma_t^2}||\mu_\theta(x_t^i,t,c^i)-\mu_\theta(x_t^i,t)||^2]+C,
\end{aligned}
\end{equation}where C is constant, and
\begin{equation}
\begin{aligned}
\mu_\theta(x_t^i,t,c^i)&=\frac{1}{\sqrt{\alpha_t}}x_t^i-\frac{\beta_t}{\sqrt{\alpha_t}\sqrt{1-\bar\alpha_t}}\epsilon_\theta(x_t^i,t,c^i), \\ 
\mu_\theta(x_t^i,t) &=\frac{1}{\sqrt{\alpha_t}}x_t^i-\frac{\beta_t}{\sqrt{\alpha_t}\sqrt{1-\bar\alpha_t}}\epsilon_\theta(x_t^i,t),
\end{aligned}
\end{equation}where $\alpha_t = 1-\beta_t, \bar{\alpha}_t\!\!=\!\!\prod_{i=1}^t (1-\beta_i)$, $\{\beta_t\}_{1:T}$ is the variance schedule. Then, $\mathcal{L}_{al}$ simplifies to:
\begin{equation}\label{eq:mse loss simplify}
\begin{aligned}
%\mathcal{L}^{i,t}_{al}=||\frac{\beta_t}{\sqrt{\alpha_t}\sqrt{1-\bar\alpha_t}}(\epsilon_\theta(x_t^i,t,c^i)-\epsilon_\theta(x_t^i,t))||^2.
\mathcal{L}^{i,t}_{al}\propto ||\epsilon_\theta(x_t^i,t,c^i)-\epsilon_\theta(x_t^i,t)||^2.
\end{aligned}
\end{equation}

 As shown in Fig.~\ref{fig:method}, the anchor vector $h_{anc}^i$ is fed into decode network $d(*)$ to get unconditional noise estimation $\epsilon_{unc}^i := \epsilon_\theta(x_t^i; t, \emptyset)$.
 Meanwhile, the anchor image $x_t^i$ is also incorporated with the class label condition $\mathbf{c^i}$ fed into the network to get the conditional noise estimation $\epsilon_{cond}^i := \epsilon_\theta(x_t^i; t, \mathbf{c^i})$.
 The alignment loss is defined between the unconditional and conditional noise estimation:
 % \footnote{Compared to CBDM loss~\cite{cbdm}, we minimize the distance between conditional generation and unconditional generation in image space, while CBDM minimizes distances between conditional generations for two different conditions. Our motivation is different from CBDM, but we end up with a similar but different loss. Our work offers different interpretations of previous work and makes insightful connections.}

 \begin{equation}\label{eq:mse loss}
 \begin{aligned}
 \mathcal{L}_{al} = \frac{1}{|B|}\sum_{i\in B}{\mathbb{E}_{t,x_0}[\frac{t}{T}\|\epsilon_\theta(x_t^i; t, \mathbf{c^i}) - \epsilon_\theta(x_t^i; t, \emptyset)\|^2}].
 \end{aligned}
 \end{equation}

The loss is weighted linearly by timesteps $t$, so that the initial steps with large $t$ are weighted more.
In other words, we align conditional generation and unconditional generation for the initial steps.% of the reverse process.
%

% \begin{figure}[t]
% \centering
% \subfigure[]{
% \includegraphics[width=1.5in]{figures/noise_1.png} 
% }
% \subfigure[]{
% \includegraphics[width=1.5in]{figures/noise_2.png} 
% }
% \DeclareGraphicsExtensions.
% \caption{\TODO{Update Fig.} Reverse processing of DDPM. 
% The first row in each figure illustrates unconditional generation, while the second to fourth rows depict conditional generation conditioned on different class labels (`dog', `cat', and `panda') by starting from the same Gaussian noise as the first row. The unconditional and conditional generation results exhibit similar object shapes.
% In Figure (a), all images display a consistent salient green region in the top-left corner during the initial time steps, as seen in the second column.
% In Figure (b), at the initial time steps (second column), the first and third-row images are difficult to distinguish, while the second and fourth-row images also exhibit similar modes.
% For the final synthesized results in the last column of Figure (b), animals of the first to third rows share the same shadow regions, while the shadow becomes part of the panda's black body in the fourth row.
% }
% \label{fig:reverse processing}
% \end{figure}

\noindent \textbf{Connection to Conditional-unconditional Alignment for Long-tailed GAN}\quad
We take inspiration from UTLO~\cite{utlo} and Transitional-GAN~\cite{transitioncgan}, which addresses long-tailed generation with GANs, and observes that the similarity between head class and tail class images increases at lower resolution representations.
To share knowledge between the head class and tail class, UTLO~\cite{utlo} proposes unconditional GAN objectives for lower-resolution representations and conditional GAN objectives for subsequent higher-resolution images.
More specifically, utilizing unconditional generation for lower resolution is effective in increasing the diversity and quality of tail class images.
%
%Transitional-GAN~\cite{transitioncgan} leverages unconditional generation in a two-staged training scheme, as unconditional generation can give a better FID than conditional generation when data is limited.

\begin{figure}[t]
\centering
\includegraphics[width=0.95\linewidth]{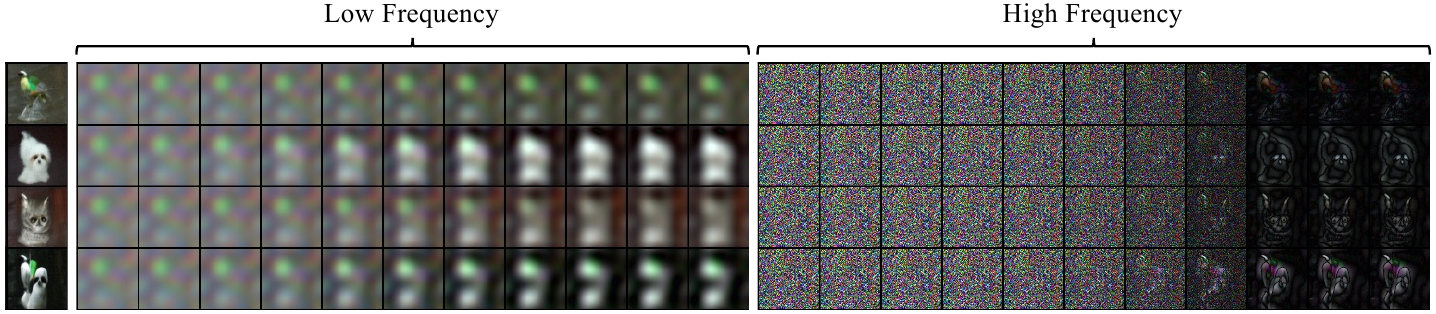}
\caption{
 The leftmost column shows synthetic images with different classes but with the same initial noise or random seed. We visualize the low-frequency component and the high-frequency component from the reverse processing.
 It shows that low-frequency components are similar during initial time steps for different classes with the same initial noise~\cite{si2024freeu}. More examples are in Appendix~\ref{sec:reverseprocessing}.}%Fig.~\ref{fig:reverse-1},~\ref{fig:reverse-2},~\ref{fig:reverse-3},~\ref{fig:reverse-4} in the Appendix}
\label{fig:reverse processing}
\end{figure}
%\textbf{\meng{Conditional-unconditinal Alignment} for Class-imbalanced Diffusion Model}\quad
%
Our alignment loss for diffusion model is similar to the \meng{conditional-unconditional alignment} approach used in GANs~\cite{utlo,transitioncgan}.
Our key insight is that diffusion models denoise images recursively, and the similarity between head class images and tail class images is higher during the \textit{initial timesteps} of the denoising process.
This is similar but different from UTLO~\cite{utlo}, which proposes unconditional generation for \textit{lower-resolution} representations.
We visualize the reverse processing of unconditional generation and conditional generation with different class labels of the original DDPM, starting from the same Gaussian noise (Fig.~\ref{fig:reverse processing}).
Images from unconditional generation and conditional generation share similar low-frequency components.
%e.g., the coarse shape of synthesized objects and background region, especially at the initial time steps\meng{~\cite{si2024freeu}}.
%
Our alignment loss matches unconditional generation and conditional generation at the initial time steps, enabling knowledge sharing between tail classes and head classes with abundant data.
%
%\meng{Another motivation is that class conditions are not necessary for all timesteps. In fact, using adaptive guidance weight with a small weight in the initial steps can yield a better FID compared to using constant guidance weight~\cite{omega_scheduler}.}

%The challenge is that there is no weight scheduler, e.g., linear scheduler, that is optimal for all datasets. So we propose MSE loss during training to make generation agnostic to class conditions for initial steps, while still using a fixed guidance weight at different timesteps during inference. Note that the guidance weight scheduler has not been explored in the context of imbalanced diffusion models.

\subsection{Overall Framework}
\label{sec:framework}

Our final loss function $\mathcal{L}$ is the sum of the standard DDPM~\cite{ddpm} loss $\mathcal{L}_{ddpm}$ and our contrastive conditional-unconditional alignment loss $\mathcal{L}_{ccua}$:
 \begin{equation}\label{eq:final loss}
 \begin{aligned}
 \mathcal{L}_{ccua} =  \alpha\cdot\mathcal{L}_{ucl} + \gamma\cdot\mathcal{L}_{al},
 \end{aligned}
 \end{equation}where
 ~$\alpha$ and $\gamma$ are the hyper-parameters for our unsupervised contrastive loss and alignment loss.
 % where $\mathcal{L}_{ddpm}$ is a standard mse loss for matching the estimated noise item with the ground-truth noise item, same as in DDPM~\cite{ddpm}.%from Eq.~\ref{eq:ddpm loss}.
 %
 The overall algorithm framework is shown as Algorithm~\ref{diffclr_algo}.
 %In classifier-free guidance (CFG), the class labels $\mathbf{c}$ are randomly dropped by a specific probability $p_{uncond}$.
 %In our setting, we keep the $p_{uncond}=10\%$, same as~\cite{cbdm}.
 % \TODO{CHECK: For unconditional training of CFG, we only update the network by the original DDPM loss $\mathcal{L}_{ddpm}$ without applying the proposed contrastive learning losses.}
 % For conditional training of CFG, for each image-class pair $(x_0^i, c^i)$ in a mini-batch $B$, we calculate the $\mathcal{L}_{ucl}$ and $\mathcal{L}_{al}$ according to Eq.~\ref{eq:nce loss} and Eq.~\ref{eq:mse loss}, respectively.
 The parameters of the noise estimation network $\epsilon_\theta$ are updated by the gradient of the final loss $\nabla_\theta\mathcal{L}$ as in Eq.~\ref{eq:final loss}.

\setlength{\intextsep}{0pt}
\begin{wrapfigure}[23]{r}{0.62\textwidth}
    % %\vspace{-0mm} % Optional: adjust vertical space
    \begin{minipage}{0.62\textwidth}
\begin{algorithm}[H]
\caption{Training algorithm of \textbf{CCUA}.}\label{diffclr_algo}
\begin{algorithmic}
    %\FOR {each batch}
        \State Set $\mathcal{L}_{ddpm}, \mathcal{L}_{ucl}, \mathcal{L}_{al} = 0$
        \For {each image-class pair $(x_0^{i}, c^{i})$ in this batch $B$}
            \State Sample $\epsilon^i \sim \mathcal{N}(\mathbf{0}, \mathbf{I})$, $t\sim \mathcal{U}(\{0, 1, ..., T\})$
            \State $x^{i}_{t} = \sqrt{\bar{\alpha}_{t}}x^{i}_0+ \sqrt{1-\bar{\alpha}_{t}}\epsilon^i$
            \State $\epsilon_{unc}^i = \epsilon_\theta(x_t^i;t,\emptyset)$,
            \State $\epsilon_{cond}^i = \epsilon_\theta(x_t^i;t,c^i)$,    \State $h_{anc}^i=e(x_t^i)$, \State $\pi_{anc} = \exp(h_{anc}^i \cdot h_{anc}^i/\tau)$, 
            \State Set $\pi_{neg} = 0$
            \For {$x_t^{j}$ in this batch with $i\neq j$}
            \State $h_{neg}^j=e(x_t^j)$
            \State $\pi_{neg} = \pi_{neg} + \exp(h_{anc}^i \cdot h_{neg}^j/\tau)$
            \EndFor
            \State $\mathcal{L}_{ucl} = \mathcal{L}_{ucl} + (-\log\frac{\pi_{anc}}{\pi_{anc}+\pi_{neg}})$

            \If {Unconditional Training of CFG}
                % \STATE \# unconditional diffusion loss
                \State $\mathcal{L}_{ddpm} = \mathcal{L}_{ddpm} + \| \epsilon^{i} - \epsilon_{unc}^i\|^2$
            \ElsIf {Conditional Training of CFG}                
                % \STATE \# conditional diffusion loss
                \State $\mathcal{L}_{ddpm} = \mathcal{L}_{ddpm} + \| \epsilon^{i} - \epsilon_{cond}^i\|^2$
                % \STATE \# MSE contrastive loss
                \State $\mathcal{L}_{al} = \mathcal{L}_{al} + \frac{t}{T}\|\epsilon_{unc}^i - \epsilon_{cond}^i\|^2$  
            \EndIf
        \EndFor
        %\State Compute the gradient with the loss
            \State $\mathcal{L}_{ccua} = \frac{1}{|B|} (\mathcal{L}_{ddpm} + \alpha\cdot\mathcal{L}_{ucl} + \gamma\cdot\mathcal{L}_{al})$
    %\ENDFOR

\end{algorithmic}
\label{alg:training}
\end{algorithm}

\end{minipage}
\end{wrapfigure}

\ICLRrebuttal{\textbf{Synergy between unsupervised contrastive loss and alignment loss} Our proposed two losses including unsupervised contrastive loss $\mathcal{L}_{ucl}$ and $\mathcal{L}_{al}$ are novel by themselves in the context of diffusion models. %
What's more, the synergy of the two losses makes our method effective.
Our $\mathcal{L}_{ucl}$ loss diversifies \textit{unconditional} latents repulsing negative pairs, while $\mathcal{L}_{al}$ aligns \textit{conditional} and \textit{unconditional} latents/representations. As a result, conditional latents are diversified implicitly.}
We choose not to directly diversify conditional latents, as is done in concurrent work of Dispersive Loss~\cite{wang2025diffuse}. Our combined loss is more effective for addressing overfitting in tail classes empirically verified in Tab.~\ref{tab: SiT on ImageNet-LT}. The geometric interpretation for the reason is that the diversity of conditional latents involves both inter-class variance and intra-class variance. Directly applying InfoNCE loss on conditional latent~\cite{wang2025diffuse} can lead to a shortcut solution that dominantly maximizes inter-class variance. However, contrastive regularization on unconditional latents \textit{must} diversity latents for all data regardless of class. Aligning conditional latents to diversified unconditional latents via distillation is more effective.

\section{Experiments}
\label{sec:exp}

\subsection{Experimental Setup}

 We report main results on ImageNet, TinyImageNet, and Places datasets, with their long-tailed version, while we also report results on CIFAR10/CIFAR100 long-tailed datasets as shown in Appendix~\ref{sec:quantative appendix}.
 We measure IS, FID, KID~\cite{kid}, spatial FID~\cite{dhariwal2021diffusion} (sFID), Precision, Recall, and FID of tail classes (FID$_{tail}$) as the evaluation metrics.
 More implementation details are provided in Appendix~\ref{sec:details}.

\begin{table}[!tb]
\caption{Comparison on ImageNet-LT~$256 \times 256$ with SiT pipeline. We use blue parentheses \textcolor{blue}{`()'} to highlight the improvement of our method over SiT baseline on FID and Recall, which denotes the overall quality and diversity, respectively. \ICLRrebuttal{Our method dramatically improves $\text{FID}_{tail}$ in particular.}}
% %\vspace{-2mm}
\label{tab: SiT on ImageNet-LT}
\begin{center}
\setlength{\tabcolsep}{0.4mm}{
\small
\begin{tabular}{c|c|c|l|c|c|c|l}
\hline
\hline

Steps & Method & IS $\uparrow$ & FID $\downarrow$ & sFID $\downarrow$ & Prec. \% $\uparrow$ & Recall \% $\uparrow$ & \ICLRrebuttal{FID$_{tail}$ $\downarrow$} \\
\hline

\multirow{5}{*}{250k} & SiT & 53.9 & 33.8 & 22.6 & 54.5 & 19.1 & \ICLRrebuttal{52.8} \\
& \ICLRrebuttal{CBDM} & \ICLRrebuttal{54.8} & \ICLRrebuttal{34.1} & \ICLRrebuttal{23.3} & \ICLRrebuttal{53.9} & \ICLRrebuttal{18.7} & \ICLRrebuttal{53.1} \\
& \ICLRrebuttal{REPA} & \ICLRrebuttal{\textbf{74.1}} & \ICLRrebuttal{28.4} & \ICLRrebuttal{\textbf{20.0}} & \ICLRrebuttal{58.3} & \ICLRrebuttal{16.1} & \ICLRrebuttal{48.5} \\
& Dispersive Loss & 53.8 & 34.0 & 22.6 & 54.8 & 19.8 & 54.5 \\
\cline{2-8}
& \textbf{CCUA (ours)} & 70.7 & \textbf{27.0} \textcolor{blue}{(-6.8)} \ & 20.5 & \textbf{60.2} & \textbf{20.1} & \textbf{37.0} \textcolor{blue}{(-15.8)} \\
% & \textbf{CCUA-N (ours)} & 73.6 & \textbf{25.8} \textcolor{blue}{(-8.0)} & \textbf{16.5} & \textbf{58.6} & \textbf{23.1} \textcolor{blue}{(+4.0)} & \ICLRrebuttal{\textbf{28.1}} \\

\hline

\multirow{5}{*}{450k} & SiT & 78.8 & 25.7 & 21.9 & 64.3 & 18.5 & \ICLRrebuttal{41.6} \\
& \ICLRrebuttal{CBDM} & \ICLRrebuttal{84.0} & \ICLRrebuttal{24.7} & \ICLRrebuttal{21.8} & \ICLRrebuttal{64.6} & \ICLRrebuttal{18.7} & \ICLRrebuttal{40.7} \\
& \ICLRrebuttal{REPA} & \ICLRrebuttal{105.9} & \ICLRrebuttal{21.9} & \ICLRrebuttal{19.5} & \ICLRrebuttal{65.7} & \ICLRrebuttal{15.2} & \ICLRrebuttal{39.7} \\
& Dispersive Loss & 83.8 & 25.2 & 21.7 & 65.5 & 17.3 & 43.0 \\
\cline{2-8}
& \ICLRrebuttal{\textbf{CCUA (ours)}} & \textbf{111.9} & \textbf{19.4} \textcolor{blue}{(-6.3)} & \textbf{18.3} & \textbf{69.4} & \textbf{18.9} & \textbf{28.8} \textcolor{blue}{(-12.8)} \\
% & \textbf{CCUA-N (ours)} & 103.6 & \textbf{19.9} \textcolor{blue}{(-5.8)} & \textbf{14.8} & \textbf{65.8} & \textbf{21.3} \textcolor{blue}{(+2.8)} & \ICLRrebuttal{\textbf{21.9}} \\

\hline

\multirow{5}{*}{700k} & SiT & 103.1 & 21.2 & 20.1 & 69.3 & 18.2 & \ICLRrebuttal{35.4} \\
& \ICLRrebuttal{CBDM} & \ICLRrebuttal{105.7} & \ICLRrebuttal{20.94} & \ICLRrebuttal{20.7} & \ICLRrebuttal{70.3} & \ICLRrebuttal{17.8} & \ICLRrebuttal{35.5} \\
& \ICLRrebuttal{REPA} & \ICLRrebuttal{126.8} & \ICLRrebuttal{19.7} & \ICLRrebuttal{20.3} & \ICLRrebuttal{68.0} & \ICLRrebuttal{15.8} & \ICLRrebuttal{36.9} \\
& Dispersive Loss & 104.0 & 21.3 & 20.4 & 68.0 & \textbf{18.5} & \ICLRrebuttal{35.9} \\
\cline{2-8}
& \ICLRrebuttal{\textbf{ CCUA (ours)\protect\footnotemark }} & \textbf{140.5} & \textbf{16.3} \textcolor{blue}{(-4.9)} & \textbf{17.2} & \textbf{73.9} & 18.4 & \textbf{24.5} \textcolor{blue}{(-10.9)} \\
% & \textbf{CCUA-N (ours)} & 119.1 & \textbf{17.5} \textcolor{blue}{(-3.7)} & \textbf{13.8} & 68.3 & \textbf{21.2} \textcolor{blue}{(+3.0)} & \ICLRrebuttal{\textbf{20.0}} \\

\hline

\multirow{5}{*}{900k} & SiT & 111.7 & 19.9 & 20.1 & 70.3 & 18.6 & \ICLRrebuttal{33.9} \\
& \ICLRrebuttal{CBDM} & \ICLRrebuttal{117.2} & \ICLRrebuttal{19.4} & \ICLRrebuttal{20.2} & \ICLRrebuttal{72.5} & \ICLRrebuttal{17.7} & \ICLRrebuttal{32.7} \\
& \ICLRrebuttal{REPA} & \ICLRrebuttal{137.8} & \ICLRrebuttal{18.1} & \ICLRrebuttal{19.3} & \ICLRrebuttal{69.8} & \ICLRrebuttal{16.2} & \ICLRrebuttal{33.8} \\
& Dispersive Loss & 115.6 & 19.7 & 20.1 & 69.9 & \textbf{18.9} & \ICLRrebuttal{34.1} \\
\cline{2-8}
& \ICLRrebuttal{\textbf{ CCUA (ours) }} & \textbf{153.1} & \textbf{15.1} \textcolor{blue}{(-4.8)} & \textbf{16.5} & \textbf{75.7} & 17.3 & \textbf{22.5} \textcolor{blue}{(-11.4)} \\
% & \textbf{CCUA (ours)} & 124.8 & \textbf{16.4} \textcolor{blue}{(-3.5)} & \textbf{13.1} & 69.8 & 21.3 \textcolor{blue}{(+2.7)} & \ICLRrebuttal{\textbf{19.5}} \\

 % \eccvfang{700k} & \ICLRrebuttal{\textbf{CCUA (ours)\protect\footnotemark}} & \ICLRrebuttal{119.0} & \ICLRrebuttal{17.2 \textcolor{blue}{(-2.7)}} & \ICLRrebuttal{13.9} & \ICLRrebuttal{68.4} & \ICLRrebuttal{\textbf{21.6}} & \ICLRrebuttal{\textbf{20.1}} \textcolor{blue}{(-13.8)} \\

% \hline

% \multirow{2}{*}{240} & \ICLRrebuttal{SiT} & \ICLRrebuttal{132.2} & \ICLRrebuttal{17.8} & \ICLRrebuttal{19.6} & \ICLRrebuttal{\textbf{74.4}} & \ICLRrebuttal{17.6} & \ICLRrebuttal{29.6} \\
% & \ICLRrebuttal{\textbf{CCUA (ours)}} & \ICLRrebuttal{\textbf{141.8}} & \ICLRrebuttal{\textbf{14.6} \textcolor{blue}{(-3.2)}} & \ICLRrebuttal{\textbf{13.2}} & \ICLRrebuttal{73.5} & \ICLRrebuttal{\textbf{19.6} \textcolor{blue}{(+2.0)}} & \ICLRrebuttal{\textbf{21.6}} \\ 

% \hline
% \hline

% \multirow{2}{*}{Uncond.} 
% & \ICLRrebuttal{SiT} & \ICLRrebuttal{9.9} & \ICLRrebuttal{109.2} & \ICLRrebuttal{15.2} & \ICLRrebuttal{21.5} & \ICLRrebuttal{37.8} \\
% & \ICLRrebuttal{CCUA} & \ICLRrebuttal{10.7} & \ICLRrebuttal{106.0} & \ICLRrebuttal{13.7} & \ICLRrebuttal{21.0} & \ICLRrebuttal{44.2} \\

\hline
\hline
\end{tabular}
}
% %\vspace{-6mm}
\end{center}
\end{table}
\footnotetext{Our method incurs longer training time as discussed in Appendix.~\ref{sec:acceleratedimplementation}. However, CCUA still outperforms SiT under the same training time (i.e., 700k for CCUA, 900k for SiT).}

\subsection{Quantitative Results}

\textbf{Class-imbalanced Generation for Diffusion Transformer}\quad
 We apply the proposed CCUA framework into SiT~\cite{ma2024sit} on ImageNet-LT dataset, and compare to CBDM~\cite{cbdm}, a long-tailed training method for UNet-based model, and REPA~\cite{repa}, the latest training technique for general DiT/SiT-based model.
 We also compare our method to the concurrent work of Dispersive Loss~\cite{wang2025diffuse} which is a contrastive loss originally developed for improving diffusion models on balanced datasets.
 % %
 % All models are sampled with ODE flow 50 steps for conditional generation with CFG strength 7.5.
 As shown in Tab.~\ref{tab: SiT on ImageNet-LT}, our method achieves remarkable improvement compared to SiT, \ICLRrebuttal{CBDM}, \ICLRrebuttal{REPA} and Dispersive Loss on various training steps.
 Our method achieves about 20\% improvement on overall FID, 30\% improvement on IS Score and FID$_{tail}$, illustrating the effectiveness of CCUA.% on long-tailed datasets.

\begin{table}[!t]
\caption{Comparison on TinyImageNet-LT $64\times64$ and Places-LT $64\times64$ with DDPM pipeline. Blue `\textcolor{blue}{()}' shows improvement of our method over DDPM baseline. Green `\textcolor{ForestGreen}{()}' shows improvement of DDPM trained on balanced version over long-tailed version.}
\label{tab: DDPM on TinyImageNet-LT and Places-LT}
% %\vspace{-3mm}
\begin{center}
\setlength{\tabcolsep}{1.6mm}{

\small
\begin{tabular}{l|l|l|l|l}
\hline
\hline
Dataset & Method & FID$\downarrow$ & FID$_{tail}\downarrow$ & KID$_{\times1k}\downarrow$ \\%& Recall$_{tail}$ \\

\hline
\multirow{5}{*}{TinyImageNet-LT}
& DDPM$^*_{bal}$~\cite{ddpm} & 15.7 \textcolor{ForestGreen}{(-3.0)} & 25.6 \textcolor{ForestGreen}{(-14.5)} & 3.2 \textcolor{ForestGreen}{(-3.1)} \\
\cline{2-5}
& DDPM~\cite{ddpm} & 18.7 & 40.1 & 6.3 \\%&0.547 \\
& CBDM~\cite{cbdm} & 20.9 & 48.1 & 6.6 \\
% & \ICLRrebuttal{CB-DDPM~\cite{cbdm}} & \ICLRrebuttal{17.41} & \ICLRrebuttal{39.03} & \ICLRrebuttal{5.67} \\
& OCLT~\cite{oclt} & 17.7 & 39.7 & 5.6 \\
\cline{2-5}
% & CCUA w/o Batch Resample & 17.16 \textcolor{blue}{(-1.51)} & 37.88 \textcolor{blue}{(-2.24)} & 4.89 \textcolor{blue}{(-1.42)} \\ %& 0.512 \\
& \textbf{CCUA (ours)} & \textbf{15.2} \textcolor{blue}{(-3.5)} & \textbf{30.4} \textcolor{blue}{(-9.7)} & \textbf{3.8} \textcolor{blue}{(-2.5)} \\

\hline
\multirow{4}{*}{Places-LT} 
% & DDIM$^*_{bal}$ & - & - & - & - \\
% \cline{2-6}
% & DDIM$^*_{bal}$ & - & - & - & - \\
% \cline{2-6}
& DDPM~\cite{ddpm} & 13.9 & 23.7 & 5.3  \\
& CBDM~\cite{cbdm} & 15.2 & 26.1 & 5.6  \\
% & \ICLRrebuttal{CB-DDPM~\cite{cbdm}} & \ICLRrebuttal{12.13} & \ICLRrebuttal{21.34} & \ICLRrebuttal{3.62} \\
& OCLT~\cite{oclt} & 13.0 & 22.8 & 4.2  \\
\cline{2-5}
& \textbf{CCUA (ours)} & \textbf{12.0} \textcolor{blue}{(-1.9)} & \textbf{20.8} \textcolor{blue}{(-2.9)} & \textbf{3.6} \textcolor{blue}{(-1.7)}  \\
% & CCUA (ours) & \\

\hline
\hline
\end{tabular}
}
% %\vspace{-6mm}
\end{center}
\end{table}

\textbf{Class-imbalanced Generation for DDPM}\quad
 For training unet-based architecture DDPM on TinyImageNet-LT and Places-LT datasets, our method also achieves the best performance, as shown in Tab.~\ref{tab: DDPM on TinyImageNet-LT and Places-LT}.
 % All models are sampled with DDIM 100 steps for conditional generation with optimal CFG strength.
 For TinyImageNet-LT datasets, we provide the metrics of the DDPM model trained on the original balanced datasets, denoted by DDPM$^*_{bal}$, as the theoretical optimal reference.
 The best FID and KID score of our method illustrates the high quality of images synthesized by our method.%\footnote{
 %The FID score of our reproduced DDPM, CBDM~\cite{cbdm}, and OCLT~\cite{oclt} are better than the numbers reported in~\cite{cbdm,oclt}. 
 %We train all methods from scratch without fine-tuning and find the optimal guidance strength $\omega$ for each method. Such implementation is different from the one used in~\cite{cbdm, oclt} but is considered more fair.}.
 % For the FID of tail classes, our method improves the FID$_{tail}$ from \textbf{40.12} to \textbf{37.88} on TinyImageNet-LT, and from \textbf{23.74} to \textbf{20.84} on Places-LT, respectively.
 % These improvements illustrate improved quality for tail class images.

We provide a comparison to other methods for long-tailed diffusion models including DiffROP~\cite{diffrop} in Appendix~\ref{sec:quantative appendix}.

\begin{table}[!t]
\caption{FID score for three `super-categories': `Head', `Body', and `Tail'.} %All models are measured with DDIM~\cite{ddim} 100 sampling steps for conditional generation with CFG.}
\label{tab:super category}
%\vspace{-3mm}
\begin{center}
\setlength{\tabcolsep}{2mm}{

\small
\begin{tabular}{l|l|c|c|c|c}
\hline
\hline
Dataset & \diagbox{Method}{FID$\downarrow$}{$P_{category}$} & \makecell[c]{Head\\$\sim80\%$} & \makecell[c]{Body\\$\sim17\%$} & \makecell[c]{Tail\\$\sim3\%$} & All \\
\hline
% \multirow{4}{*}{CIFAR100LT \TODO{Replaced by Places-LT}}
%  & DDIM & 8.75 & 10.88 & 17.45 & 6.95 \\
%  & CBDM & 8.19 & 10.54 & 17.36 & 6.50 \\
%  & OCLT & 8.37  & 11.44 & 17.22 & 6.45 \\
% \cline{2-6}
%  & \textbf{CLDM (Ours)} & \textbf{7.53} & \textbf{10.19} & \textbf{16.35} & \textbf{6.24} \\
%  % \textbf{Ours mse1.0 nce1.0} & 7.53 & 10.19 & 16.35 & 6.24 \\

\hline
% \multirow{2}{*}{ImageNet-LT-40 Epochs}
%  & SiT~\cite{ma2024sit} & 23.39 & - & 52.82 \\
%  & \textbf{CCUA (ours)} & 21.58 & - & 28.11 \\ 
% \multirow{2}{*}{ImageNet-LT-80 Epochs}
%  & SiT~\cite{ma2024sit} & 18.54 & - & 41.61 \\
%  & \textbf{CCUA (ours)} & 17.18 & - & 21.97 \\ 
% \multirow{2}{*}{ImageNet-LT-120 Epochs}
%  & SiT~\cite{ma2024sit} & 17.54 & - & 35.48 \\
%  & \textbf{CCUA (ours)} & 17.45 & - & 20.00 \\ 
% \multirow{2}{*}{ImageNet-LT-160 Epochs}
%  & SiT~\cite{ma2024sit} & 17.88 & - & 33.92 \\
%  & \textbf{CCUA (ours)} & 17.86 & - & 19.55 \\ 
%  \hline

% \multirow{4}{*}{TinyImageNet-LT}
%  & DDPM~\cite{ddpm} & 21.27 & 33.25 & 40.12 & 18.67 \\
%  & CBDM~\cite{cbdm} & 24.12 & 35.02 & 48.07 & 20.90 \\
%  & OCLT~\cite{oclt} & \textbf{20.64}  & 30.62 & 39.67 & 17.72 \\
% \cline{2-6}
%  % & \textbf{CCUA (ours)} & \textbf{20.39} & 31.02 & \textbf{37.88} & \textbf{17.16} \\
%  & \textbf{CCUA (ours)} & 21.32 & \textbf{27.98} & \textbf{30.39} & \textbf{15.24} \\

% \hline
% \multirow{4}{*}{Places-LT}
%  & DDPM~\cite{ddpm} & 19.26 & 20.31 & 23.74 & 13.89 \\
%  & CBDM~\cite{cbdm} & 21.26 & 21.68 & 26.07 & 15.15 \\
%  & OCLT~\cite{oclt} & 19.16 & 19.43 & 22.84 & 13.04 \\
% \cline{2-6}
%  & \textbf{CCUA (ours)} & \textbf{18.15} & \textbf{19.27} & \textbf{20.84} & \textbf{11.99} \\
%  % \textbf{Ours mse1.0 nce1.0} & 7.53 & 10.19 & 16.35 & 6.24 \\

\multirow{4}{*}{TinyImageNet-LT}
 & DDPM~\cite{ddpm} & 21.3 & 33.3 & 40.1 & 18.7 \\
 & CBDM~\cite{cbdm} & 24.1 & 35.0 & 48.1 & 20.9 \\
 & OCLT~\cite{oclt} & \textbf{20.6}  & 30.6 & 39.7 & 17.7 \\
\cline{2-6}
 % & \textbf{CCUA (ours)} & \textbf{20.39} & 31.02 & \textbf{37.88} & \textbf{17.16} \\
 & \textbf{CCUA (ours)} & 21.3 & \textbf{28.0} & \textbf{30.4} & \textbf{15.2} \\

\hline
\multirow{4}{*}{Places-LT}
 & DDPM~\cite{ddpm} & 19.3 & 20.3 & 23.7 & 13.9 \\
 & CBDM~\cite{cbdm} & 21.3 & 21.7 & 26.1 & 15.2 \\
 & OCLT~\cite{oclt} & 19.2 & 19.4 & 22.8 & 13.0 \\
\cline{2-6}
 & \textbf{CCUA (ours)} & \textbf{18.2} & \textbf{19.3} & \textbf{20.8} & \textbf{12.0} \\
 % \textbf{Ours mse1.0 nce1.0} & 7.53 & 10.19 & 16.35 & 6.24 \\

\hline
\hline
\end{tabular}
}
\end{center}
\end{table}

\textbf{Results Across Various Category Intervals}\quad
 To see the effectiveness of our method on tail classes, we divide each dataset into three super categories: `Head', `Body', and `Tail', with classes sorted by the number of training images.
 %Same as~\cite{diffrop}, each dataset was divided by sorting classes in descending order of size and distributing them equally.
 %For CIFAR10LT dataset, the top 3 classes were assigned to the “Head” category, the next 4 to the “Body” category, and the remaining 3 to the “Tail” category.
 For each dataset, the top 33\% classes were allocated to the `Head' category, the next 34\% classes to the `Body' category, and the rest to the `Tail' category.
 The percentage $P_{category}$ of training images belonging to each category is shown in the header of Tab.~\ref{tab:super category}.
 Our method achieves better FID score for `Body' and `Tail' categories, while keeps `Head' category unchanged or even better.

\subsection{Qualitative Results and Ablation Study}
\label{sec:visualization and analysis}

\begin{figure}[!t]
\centering
\begin{minipage}{0.7\columnwidth}
    \subfloat[Synthetic images from SiT~\cite{ma2024sit}.]{
      \label{sit_vis}
      \includegraphics[width=0.98\columnwidth]{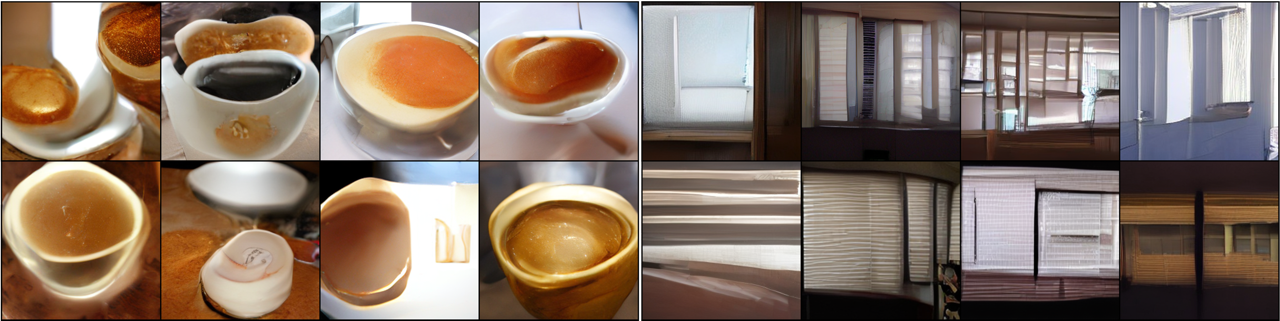}
      }
\end{minipage}
\begin{minipage}{0.7\columnwidth}
    \subfloat[Synthetic images from CCUA (ours).]{
      \label{ccua_vis}
        \includegraphics[width=0.98\columnwidth]{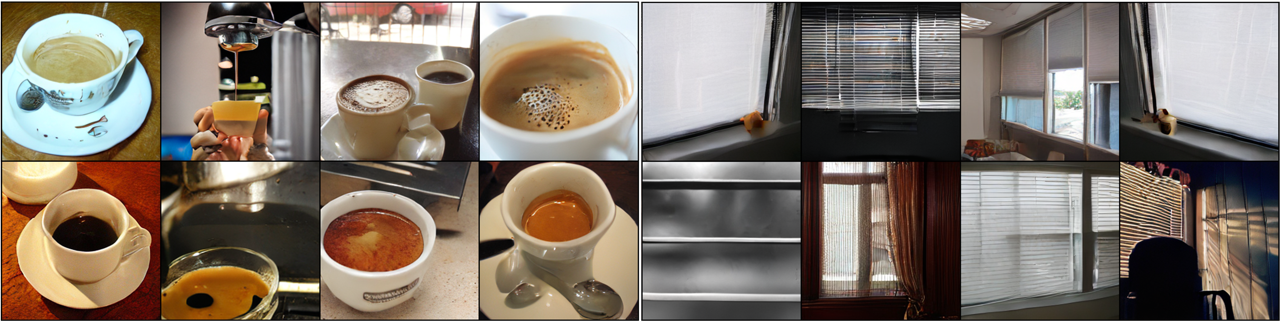}
      }
\end{minipage}

\caption{Synthetic images of SiT and it with CCUA for ImageNet-LT tail classes `espresso' and `window shade'.
All methods start the denoising process from the same Gaussian noise at corresponding grid cells.
CCUA shows consistently better diversity and fidelity.}
% \TODO{Visualization, also show training images? Meng: The images are too small. Show fewer images}
\label{fig:visualization}
\end{figure}

 \textbf{Visualization of Synthetic Images}\quad Fig.~\ref{fig:visualization} shows synthetic images of SiT and with CCUA for ImageNet-LT tail classes `espresso' and `window shade'.
 Compared to SiT, CCUA achieves visually higher fidelity and diversity.
 More qualitative results on ImageNet-LT, TinyImageNet-LT, and CIFAR100-LT are shown in Appendix~\ref{sec:qualitative appendix}.

\begin{table}[!t]
% %\vspace{-3mm}
\caption{\ICLRrebuttal{Batch Resample Strategy Analysis on TinyImageNet-LT. Blue `\textcolor{blue}{()}' highlights the improvement of each method compared to the DDPM baseline.}}
% %\vspace{-3mm}
\label{tab: batch resample tinyimagenetlt}
\begin{center}
\setlength{\tabcolsep}{5mm}{
\footnotesize
\begin{tabular}{l|c|c|c}
\hline
\hline

 Method & FID $\downarrow$ & FID$_{tail}\downarrow$ & KID$\downarrow$ \\
\hline

DDPM (baseline) & 18.7 & 40.1 & 6.3 \\
\quad+Batch Resample & 17.2~\textcolor{blue}{(-1.5)} & 33.6~\textcolor{blue}{(-6.5)} & 4.7~\textcolor{blue}{(-1.6)} \\
\hline
$\mathcal{L}_{ccua}$ (Eq.~\ref{eq:final loss}) & 17.2 \textcolor{blue}{(-1.5)} & 37.9 \textcolor{blue}{(-2.2)} & 4.9 \textcolor{blue}{(-1.4)} \\
\quad+Batch Resample & \textbf{15.2} \textcolor{blue}{(-3.5)} & \textbf{30.4} \textcolor{blue}{(-9.7)} & \textbf{3.8} \textcolor{blue}{(-2.5)} \\

\hline
\hline
\end{tabular}
}
\end{center}
% %\vspace{-10mm}
\end{table}

 \textbf{Ablation Study of Batch Resample Strategy}\quad \eccvfang{We conduct ablation study of batch resample strategy on both TinyImageNet-LT and ImageNet-LT datasets. As shown in Tab.~\ref{tab: batch resample tinyimagenetlt}, applying $\mathcal{L}_{ccua}$ itself only is already comparable to DDPM with batch resampling replacement.
 While with applying batch resample strategy, the proposed $\mathcal{L}_{ccua}$ loss benefits more from it and outperforms `DDPM + Batch Resample' on all three metrics.
 The ablation study of Batch Resample Strategy on ImageNet-LT is reported in Tab.~\ref{tab: batch resample strategy analysis} in Appendix.~\ref{sec:ablation_study_appendix}.}
 %On ImageNet-LT, as shown in Table~\ref{tab: batch resample strategy analysis} in Appendix.~\ref{sec:ablation_study_appendix}, $\mathcal{L}_{ccua}$ consistently outperforms SiT with and without applying batch resampling strategy, separately.}
 % \fang{As shown in Table~\ref{tab: batch resample strategy analysis}, we measure the performance of SiT and CCUA, with or without applying batch resampling strategy.
 % % For SiT, the batch resampling strategy leads to higher fidelity, with IS from \textbf{53.93} to \textbf{57.79}, and FID from \textbf{33.87} to \textbf{28.05}, while the declined Recall from \textbf{19.17} to \textbf{17.26} denotes the limited diversity caused by batch resampling strategy.
 % Compared to SiT, the proposed CCUA loss achieves better IS, FID, sFID and Recall with only \textbf{0.09} Precision's decline, illustrates our method on improving diversity without trading off fidelity.
 % With applying batch resample strategy, the CCUA further improves all metrics.
 % Notes that the CCUA increases Recall from \textbf{20.66} to \textbf{23.05}, while SiT suffers a decline in Recall from \textbf{19.17} to \textbf{17.26} when applying batch resampling.
 % As we discuss in Section.~\ref{sec:ucl}, the $\mathcal{L}_{ucl}$ could benefit from batch resample strategy due to the increasing instances of tail classes, leading to more diverse distribution in the latent space.
 % We also report batch resample ablation study on TinyImageNet-LT in Table~\ref{tab: batch resample tinyimagenetlt}.}

 \ICLRrebuttal{\textbf{Ablation Study of Hyper-parameters and Losses}\quad Tab.~\ref{tab: hyperparameters analysis} shows the ablation study of the hyper-parameters for the proposed $\mathcal{L}_{ucl}$ and $\mathcal{L}_{al}$ losses, which is conducted on ImageNet-LT dataset. 
 % We found that extremely large $\alpha=1$ and $\gamma=1$ would cause much higher Recall and lower Precision, which means the generated images possess better diversity but worse fidelity. 
 \eccvfang{The `Latent' denotes the latent representation $h$ used for $\mathcal{L}_{ucl}$ is from the $N/4$-th or $N$-th block for a SiT model with $N$ SiT/DiT blocks.}
 We selected the optimal $\alpha=0.05$, $\gamma=0.05$ for the best FID and IS.
 We provide more detailed ablation study about transformer block applied to $\mathcal{L}_{ccua}$ shown in Tab.~\ref{tab: latent encoder} in Appendix.~\ref{sec:ablation_study_appendix}.}

\begin{table}[!t]
% \caption{\ICLRrebuttal{Hyper-parameters Analysis of CCUA on ImageNet-LT~$256 \times 256$ with DiT-based models. All models are trained from scratch to 450k steps w/o applying Batch Resampling.}}
\caption{Hyper-parameters Analysis of CCUA on ImageNet-LT~$256 \times 256$ with DiT-based models. All models are trained from scratch to 250k steps.}
% %\vspace{-2mm}
\label{tab: hyperparameters analysis}
\begin{center}
\setlength{\tabcolsep}{1mm}{
\begin{threeparttable}
\footnotesize
\begin{tabular}{l|c|c|c|c|c|c|c|c}
\hline
\hline

Method & $\alpha$ & $\gamma$ & Latent & IS $\uparrow$ & FID $\downarrow$ & sFID $\downarrow$ & Prec. (\%) $\uparrow$ & Recall (\%) $\uparrow$ \\

\hline
SiT (baseline) & 0 & 0 & - & 53.9 & 33.8 & 22.6 & 54.5 & 19.1 \\
\hline
% \multirow{6}{*}{+CLDM (ours)} 
% & 1.0 & 1.0 & 70.38 & 25.79 & \textbf{17.48} & 56.65 & \textbf{28.39} \\
% & 0.1 & 0.1 & 84.55 & 24.12 & 20.60 & 63.83 & 20.39 \\
% & 0.05 & 0.05 & \textbf{86.28} & \textbf{23.69} & 20.56 & \textbf{65.15} & 20.21 \\
% & 0.01 & 0.01 & 83.50 & 24.63 & 21.57 & 64.87 & 19.74 \\
% \cline{2-8}
% & 0.1 & 0 & 84.69 & 24.83 & 21.93 & 64.28 & 18.70 \\
% & 0 & 0.1 & 83.59 & 24.45 & 21.12 & 63.35 & 20.16 \\
\multirow{15}{*}{\textbf{$\mathcal{L}_{ccua}$ (ours)}} 
& 1.0 & 1.0  & N/4 & 58.2 & 30.5 & \textbf{16.2} & 51.1 & \textbf{29.4} \\
& 0.1 & 0.1 & N/4 & 70.3 & 27.3 & 18.8 & 59.2 & 21.3 \\
& 0.1 & 0 & N/4 & 67.40 & 28.76 & 25.48 & 58.52 & 19.30 \\
& 0 & 0.1 & N/4 & 64.49 & 29.63 & 24.81 & 56.40 & 20.55 \\
& 0.05 & 0.05 & N/4 & \textbf{70.7} & \textbf{27.0} & 20.5 & 60.2 & 20.1 \\
% \cline{2-9}
% & 1.0 & 0 & 72.05 & 26.79 & 19.30 & 60.13 & 19.48 \\
& 0.05 & 0 & N/4 & 69.2 & 27.7 & 19.9 & 59.4 & 20.6 \\
% & 0 & 1.0 & 54.37 & 32.15 & 17.13 & 50.02 & 30.40 \\
& 0 & 0.05 & N/4 & 68.4 & 28.2 & 19.5 & 58.8 & 21.6 \\
& 0.01 & 0.01 & N/4 & 69.7 & 27.6 & 19.3 & \textbf{60.7} & 19.9 \\

% \cmidrule{2-9}\morecmidrules\cmidrule{2-9}
\cline{2-9}

& 1.0 & 1.0 & N & 52.5 & 33.0 & \textbf{16.3} & 48.9 & \textbf{29.9} \\
% & 0.5 & 0.5 & N & 62.3 & 30.1 & 20.3 & 55.6 & 22.3 \\
& 0.1 & 0.1 & N & 65.3 & 29.5 & 25.5 & 56.3 & 20.6 \\
& 0.05 & 0.05 & N & \textbf{73.6} & \textbf{25.9} & 16.5 & 58.6 & 23.1 \\
& 0.05 & 0 & N & 64.3 & 28.7 & 18.7 & 57.5 & 21.3 \\
& 0 & 0.05 & N & 64.7 & 29.0 & 20.2 & \textbf{58.6} & 20.0 \\
% & 0.05 & 0.05 & \textbf{103.66} & \textbf{19.99} & \textbf{14.88} & \textbf{65.80} & 21.31 \\
& 0.01 & 0.01 & N & 65.8 & 28.6 & 19.0 & 58.5 & 19.8 \\
% \cline{2-9}

% \hline
% SiT~\cite{ma2024sit} & - & - & 78.88 & 25.72 & 21.95 & 64.35 & 18.58 \\
% \hline
% \multirow{6}{*}{$\mathcal{L}_{ccua}$} 
% & 1.0 & 1.0 & 70.38 & 25.79 & \textbf{17.48} & 56.65 & \textbf{28.39} \\
% & 0.1 & 0.1 & 84.55 & 24.12 & 20.60 & 63.83 & 20.39 \\
% & 0.05 & 0.05 & \textbf{86.28} & \textbf{23.69} & 20.56 & \textbf{65.15} & 20.21 \\
% & 0.01 & 0.01 & 83.50 & 24.63 & 21.57 & 64.87 & 19.74 \\
% \cline{2-8}
% & 0.1 & 0 & 84.69 & 24.83 & 21.93 & 64.28 & 18.70 \\
% & 0 & 0.1 & 83.59 & 24.45 & 21.12 & 63.35 & 20.16 \\

\hline
\hline
\end{tabular}
    % \footnotesize{
    
    % }
\end{threeparttable}
}
\end{center}
\end{table}

\section{Conclusion}
\label{sec:conclusion}
Real-world data for training image generation models often exhibit long-tailed distributions.
Similar to class-imbalanced GANs~\cite{utlo}, class-imbalanced diffusion models generates inferior tail class images due to data scarcity.
% %
% Inspired by the success and impact of contrastive learning for long-tailed recognition, we are interested in whether contrastive learning can improve class-imbalanced diffusion models.
% %
% We are also motivated by conditional-unconditional alignment for class-imblanced GAN.
%With well-motivated losses and carefully crafted positive/negative pairs, we effectively leveraged contrastive learning with noticeable improvement.
%
We propose a framework with two losses that are synergistic.
Firstly, we align class-conditional generation with unconditional generation for large timesteps using an alignment loss.
This encourages the initial denoising steps to be class-agnostic, thereby enriching tail classes through knowledge sharing from head classes--a principle demonstrated to enhance long-tailed GAN performance~\cite{transitioncgan,utlo}, which we successfully adapt to diffusion models.
Secondly, we diversify unconditional generation via an unsupervised contrastive loss with negative samples only to contrast the latents of different synthetic images, promoting intra-class diversity in particular for tail classes.
%
%
%This second loss aligns conditional generation with unconditional generation for initial timesteps, which enriches tail class images with abundant data and variation presented in head class images.
%
%Our intuition is that head class and tail class images can have similar low-frequency image components in the initial steps during the denoising process.
%
With the two losses, our framework of contrastive conditional-unconditional alignment boosts the performance of DDPM~\cite{ddpm} and SiT~\cite{ma2024sit} for long-tailed image generation and outperforms alternative methods for long-tailed generation including CBDM~\cite{cbdm}, OCLT~\cite{oclt}, and concurrent work including Dispersive Loss~\cite{wang2025diffuse}.
%
%Contrastive losses have been utilized as guidance during inference for diffusion models to improve adversarial robustness~\cite{ouyang2023improving}, but we are the first to adapt contrastive learning for class-imbalanced diffusion models.
%
Extensive experiments on multiple datasets in particular ImageNet-LT with 256x256 resolution demonstrated the effectiveness of our method on both UNet-based architecture and Diffusion Transformer.

% \clearpage  % TODO FINAL: This \clearpage needs to be removed from both review and camera-ready versions.

% \section*{Acknowledgements}
% Please insert your acknowledgments here.

\bibliographystyle{splncs04}
\bibliography{main}

@String(CVPR  = {IEEE Conf. Comput. Vis. Pattern Recog.})

@String(ICML  = {Int. Conf. Mach. Learn.})

@String(ICLR  = {Int. Conf. Learn. Represent.})

@String(CVPR  = {CVPR})

@String(ICML  = {ICML})

@String(ICLR  = {ICLR})

@article{ddpm,
  title={Denoising diffusion probabilistic models},
  author={Ho, Jonathan and Jain, Ajay and Abbeel, Pieter},
  journal={Advances in neural information processing systems},
  volume={33},
  pages={6840--6851},
  year={2020}
}

@article{ddim,
  title={Denoising diffusion implicit models},
  author={Song, Jiaming and Meng, Chenlin and Ermon, Stefano},
  journal={arXiv preprint arXiv:2010.02502},
  year={2020}
}

@inproceedings{stable_diffusion,
  title={High-resolution image synthesis with latent diffusion models},
  author={Rombach, Robin and Blattmann, Andreas and Lorenz, Dominik and Esser, Patrick and Ommer, Bj{\"o}rn},
  booktitle={Proceedings of the IEEE/CVF conference on computer vision and pattern recognition},
  pages={10684--10695},
  year={2022}
}

@inproceedings{zhang2023adding,
  title={Adding conditional control to text-to-image diffusion models},
  author={Zhang, Lvmin and Rao, Anyi and Agrawala, Maneesh},
  booktitle={Proceedings of the IEEE/CVF International Conference on Computer Vision},
  pages={3836--3847},
  year={2023}
}

@inproceedings{ruiz2023dreambooth,
  title={Dreambooth: Fine tuning text-to-image diffusion models for subject-driven generation},
  author={Ruiz, Nataniel and Li, Yuanzhen and Jampani, Varun and Pritch, Yael and Rubinstein, Michael and Aberman, Kfir},
  booktitle={Proceedings of the IEEE/CVF conference on computer vision and pattern recognition},
  pages={22500--22510},
  year={2023}
}

@article{couairon2022diffedit,
  title={Diffedit: Diffusion-based semantic image editing with mask guidance},
  author={Couairon, Guillaume and Verbeek, Jakob and Schwenk, Holger and Cord, Matthieu},
  journal={arXiv preprint arXiv:2210.11427},
  year={2022}
}

@inproceedings{kawar2023imagic,
  title={Imagic: Text-based real image editing with diffusion models},
  author={Kawar, Bahjat and Zada, Shiran and Lang, Oran and Tov, Omer and Chang, Huiwen and Dekel, Tali and Mosseri, Inbar and Irani, Michal},
  booktitle={Proceedings of the IEEE/CVF Conference on Computer Vision and Pattern Recognition},
  pages={6007--6017},
  year={2023}
}

@article{ho2022imagen,
  title={Imagen video: High definition video generation with diffusion models},
  author={Ho, Jonathan and Chan, William and Saharia, Chitwan and Whang, Jay and Gao, Ruiqi and Gritsenko, Alexey and Kingma, Diederik P and Poole, Ben and Norouzi, Mohammad and Fleet, David J and others},
  journal={arXiv preprint arXiv:2210.02303},
  year={2022}
}

@article{ho2022video,
  title={Video diffusion models},
  author={Ho, Jonathan and Salimans, Tim and Gritsenko, Alexey and Chan, William and Norouzi, Mohammad and Fleet, David J},
  journal={Advances in Neural Information Processing Systems},
  volume={35},
  pages={8633--8646},
  year={2022}
}

@article{cifarlt,
  title={Learning imbalanced datasets with label-distribution-aware margin loss},
  author={Cao, Kaidi and Wei, Colin and Gaidon, Adrien and Arechiga, Nikos and Ma, Tengyu},
  journal={Advances in neural information processing systems},
  volume={32},
  year={2019}
}

@article{fid,
  title={Gans trained by a two time-scale update rule converge to a local nash equilibrium},
  author={Heusel, Martin and Ramsauer, Hubert and Unterthiner, Thomas and Nessler, Bernhard and Hochreiter, Sepp},
  journal={Advances in neural information processing systems},
  volume={30},
  year={2017}
}

@article{recall,
  title={Improved precision and recall metric for assessing generative models},
  author={Kynk{\"a}{\"a}nniemi, Tuomas and Karras, Tero and Laine, Samuli and Lehtinen, Jaakko and Aila, Timo},
  journal={Advances in neural information processing systems},
  volume={32},
  year={2019}
}

@article{cfg,
  title={Classifier-free diffusion guidance},
  author={Ho, Jonathan and Salimans, Tim},
  journal={arXiv preprint arXiv:2207.12598},
  year={2022}
}

@article{infonce,
  title={Representation learning with contrastive predictive coding},
  author={Oord, Aaron van den and Li, Yazhe and Vinyals, Oriol},
  journal={arXiv preprint arXiv:1807.03748},
  year={2018}
}

@inproceedings{cbdm,
  title={Class-balancing diffusion models},
  author={Qin, Yiming and Zheng, Huangjie and Yao, Jiangchao and Zhou, Mingyuan and Zhang, Ya},
  booktitle={Proceedings of the IEEE/CVF Conference on Computer Vision and Pattern Recognition},
  pages={18434--18443},
  year={2023}
}

@article{diffrop,
  title={Training Class-Imbalanced Diffusion Model Via Overlap Optimization},
  author={Yan, Divin and Qi, Lu and Hu, Vincent Tao and Yang, Ming-Hsuan and Tang, Meng},
  journal={arXiv preprint arXiv:2402.10821},
  year={2024}
}

@inproceedings{oclt,
  title={Long-tailed diffusion models with oriented calibration},
  author={Zhang, Tianjiao and Zheng, Huangjie and Yao, Jiangchao and Wang, Xiangfeng and Zhou, Mingyuan and Zhang, Ya and Wang, Yanfeng},
  booktitle={The Twelfth International Conference on Learning Representations},
  year={2024}
}

@inproceedings{utlo,
  title={Taming the Tail in Class-Conditional GANs: Knowledge Sharing via Unconditional Training at Lower Resolutions},
  author={Khorram, Saeed and Jiang, Mingqi and Shahbazi, Mohamad and Danesh, Mohamad H and Fuxin, Li},
  booktitle={Proceedings of the IEEE/CVF Conference on Computer Vision and Pattern Recognition},
  pages={7580--7590},
  year={2024}
}

@article{poole2022dreamfusion,
  author = {Poole, Ben and Jain, Ajay and Barron, Jonathan T. and Mildenhall, Ben},
  title = {DreamFusion: Text-to-3D using 2D Diffusion},
  journal = {arXiv},
  year = {2022},
}

@article{liu2022open,
  title={Open long-tailed recognition in a dynamic world},
  author={Liu, Ziwei and Miao, Zhongqi and Zhan, Xiaohang and Wang, Jiayun and Gong, Boqing and Stella, X Yu},
  journal={IEEE Transactions on Pattern Analysis and Machine Intelligence},
  year={2022},
  publisher={IEEE}
}

@article{tan2020fairgen,
  title   = {Improving the Fairness of Deep Generative Models without Retraining},
  author  = {Tan, Shuhan and Shen, Yujun and Zhou, Bolei},
  journal = {arXiv preprint arXiv:2012.04842},
  year    = {2020}
}

@inproceedings{jiang2021self,
  title={Self-damaging contrastive learning},
  author={Jiang, Ziyu and Chen, Tianlong and Mortazavi, Bobak J and Wang, Zhangyang},
  booktitle=ICML,
  year={2021},
}

@inproceedings{chen2020simple,
  title={A simple framework for contrastive learning of visual representations},
  author={Chen, Ting and Kornblith, Simon and Norouzi, Mohammad and Hinton, Geoffrey},
  booktitle=ICML,
  year={2020},
}

@article{khosla2020supervised,
  title={Supervised contrastive learning},
  author={Khosla, Prannay and Teterwak, Piotr and Wang, Chen and Sarna, Aaron and Tian, Yonglong and Isola, Phillip and Maschinot, Aaron and Liu, Ce and Krishnan, Dilip},
  journal={Advances in neural information processing systems},
  volume={33},
  pages={18661--18673},
  year={2020}
}

@inproceedings{ouyang2023improving,
  title={Improving adversarial robustness through the contrastive-guided diffusion process},
  author={Ouyang, Yidong and Xie, Liyan and Cheng, Guang},
  booktitle={International Conference on Machine Learning},
  pages={26699--26723},
  year={2023},
  organization={PMLR}
}

@inproceedings{szegedy2016rethinking,
  title={Rethinking the inception architecture for computer vision},
  author={Szegedy, Christian and Vanhoucke, Vincent and Ioffe, Sergey and Shlens, Jon and Wojna, Zbigniew},
  booktitle={Proceedings of the IEEE conference on computer vision and pattern recognition},
  pages={2818--2826},
  year={2016}
}

@article{tsne,
  title={Visualizing data using t-SNE.},
  author={Van der Maaten, Laurens and Hinton, Geoffrey},
  journal={Journal of machine learning research},
  volume={9},
  number={11},
  year={2008}
}

@inproceedings{rangwani2022improving,
  title={Improving gans for long-tailed data through group spectral regularization},
  author={Rangwani, Harsh and Jaswani, Naman and Karmali, Tejan and Jampani, Varun and Babu, R Venkatesh},
  booktitle={European Conference on Computer Vision},
  pages={426--442},
  year={2022},
  organization={Springer}
}

@inproceedings{rangwani2023noisytwins,
  title={Noisytwins: Class-consistent and diverse image generation through stylegans},
  author={Rangwani, Harsh and Bansal, Lavish and Sharma, Kartik and Karmali, Tejan and Jampani, Varun and Babu, R Venkatesh},
  booktitle={Proceedings of the IEEE/CVF Conference on Computer Vision and Pattern Recognition},
  pages={5987--5996},
  year={2023}
}

@article{vae,
  title={An introduction to variational autoencoders},
  author={Kingma, Diederik P and Welling, Max and others},
  journal={Foundations and Trends{\textregistered} in Machine Learning},
  volume={12},
  number={4},
  pages={307--392},
  year={2019},
  publisher={Now Publishers, Inc.}
}

@inproceedings{stylegan,
  title={A style-based generator architecture for generative adversarial networks},
  author={Karras, Tero and Laine, Samuli and Aila, Timo},
  booktitle=CVPR,
  year={2019}
}

@inproceedings{biggan,
  title={Large scale GAN training for high fidelity natural image synthesis},
  author={Brock, Andrew and Donahue, Jeff and Simonyan, Karen},
  booktitle=ICLR,
  year={2019}
}

@inproceedings{ai2023generative,
  title={Generative Oversampling for Imbalanced Data via Majority-Guided VAE},
  author={Ai, Qingzhong and Wang, Pengyun and He, Lirong and Wen, Liangjian and Pan, Lujia and Xu, Zenglin},
  booktitle={International Conference on Artificial Intelligence and Statistics},
  pages={3315--3330},
  year={2023},
  organization={PMLR}
}

@inproceedings{moco,
  title={Momentum contrast for unsupervised visual representation learning},
  author={He, Kaiming and Fan, Haoqi and Wu, Yuxin and Xie, Saining and Girshick, Ross},
  booktitle=CVPR,
  year={2020}
}

@inproceedings{li2022targeted,
  title={Targeted supervised contrastive learning for long-tailed recognition},
  author={Li, Tianhong and Cao, Peng and Yuan, Yuan and Fan, Lijie and Yang, Yuzhe and Feris, Rogerio S and Indyk, Piotr and Katabi, Dina},
  booktitle={Proceedings of the IEEE/CVF Conference on Computer Vision and Pattern Recognition},
  pages={6918--6928},
  year={2022}
}

@article{fuest2024diffusion,
  title={Diffusion models and representation learning: A survey},
  author={Fuest, Michael and Ma, Pingchuan and Gui, Ming and Fischer, Johannes S and Hu, Vincent Tao and Ommer, Bjorn},
  journal={arXiv preprint arXiv:2407.00783},
  year={2024}
}

@article{baranchuk2021label,
  title={Label-efficient semantic segmentation with diffusion models},
  author={Baranchuk, Dmitry and Rubachev, Ivan and Voynov, Andrey and Khrulkov, Valentin and Babenko, Artem},
  journal={arXiv preprint arXiv:2112.03126},
  year={2021}
}

@article{kid,
  title={Demystifying mmd gans},
  author={Bi{\'n}kowski, Miko{\l}aj and Sutherland, Danica J and Arbel, Michael and Gretton, Arthur},
  journal={arXiv preprint arXiv:1801.01401},
  year={2018}
}

@article{karras2020training,
  title={Training generative adversarial networks with limited data},
  author={Karras, Tero and Aittala, Miika and Hellsten, Janne and Laine, Samuli and Lehtinen, Jaakko and Aila, Timo},
  journal={Advances in neural information processing systems},
  volume={33},
  pages={12104--12114},
  year={2020}
}

@article{omega_scheduler,
  title={Analysis of classifier-free guidance weight schedulers},
  author={Wang, Xi and Dufour, Nicolas and Andreou, Nefeli and Cani, Marie-Paule and Abrevaya, Victoria Fern{\'a}ndez and Picard, David and Kalogeiton, Vicky},
  journal={arXiv preprint arXiv:2404.13040},
  year={2024}
}

@inproceedings{
transitioncgan,
title={Collapse by Conditioning: Training Class-conditional {GAN}s with Limited Data},
author={Mohamad Shahbazi and Martin Danelljan and Danda Pani Paudel and Luc Van Gool},
booktitle={International Conference on Learning Representations},
year={2022},
url={https://openreview.net/forum?id=7TZeCsNOUB_}
}

@article{tinyimagenet,
  title={Tiny imagenet visual recognition challenge},
  author={Le, Yann and Yang, Xuan},
  journal={CS 231N},
  volume={7},
  number={7},
  pages={3},
  year={2015}
}

@inproceedings{dalva2024noiseclr,
  title={Noiseclr: A contrastive learning approach for unsupervised discovery of interpretable directions in diffusion models},
  author={Dalva, Yusuf and Yanardag, Pinar},
  booktitle={Proceedings of the IEEE/CVF Conference on Computer Vision and Pattern Recognition},
  pages={24209--24218},
  year={2024}
}

@inproceedings{si2024freeu,
  title={Freeu: Free lunch in diffusion u-net},
  author={Si, Chenyang and Huang, Ziqi and Jiang, Yuming and Liu, Ziwei},
  booktitle={Proceedings of the IEEE/CVF Conference on Computer Vision and Pattern Recognition},
  pages={4733--4743},
  year={2024}
}

@inproceedings{ma2024sit,
  title={Sit: Exploring flow and diffusion-based generative models with scalable interpolant transformers},
  author={Ma, Nanye and Goldstein, Mark and Albergo, Michael S and Boffi, Nicholas M and Vanden-Eijnden, Eric and Xie, Saining},
  booktitle={European Conference on Computer Vision},
  pages={23--40},
  year={2024},
  organization={Springer}
}

@article{wang2025diffuse,
  title={Diffuse and Disperse: Image Generation with Representation Regularization},
  author={Wang, Runqian and He, Kaiming},
  journal={arXiv preprint arXiv:2506.09027},
  month={July},
  year={2025}
}

@article{dhariwal2021diffusion,
  title={Diffusion models beat gans on image synthesis},
  author={Dhariwal, Prafulla and Nichol, Alexander},
  journal={Advances in neural information processing systems},
  volume={34},
  pages={8780--8794},
  year={2021}
}

@article{repa,
  title={Representation alignment for generation: Training diffusion transformers is easier than you think},
  author={Yu, Sihyun and Kwak, Sangkyung and Jang, Huiwon and Jeong, Jongheon and Huang, Jonathan and Shin, Jinwoo and Xie, Saining},
  journal={arXiv preprint arXiv:2410.06940},
  year={2024}
}

@article{zhang2019balance,
  title={To balance or not to balance: A simple-yet-effective approach for learning with long-tailed distributions},
  author={Zhang, Junjie and Liu, Lingqiao and Wang, Peng and Shen, Chunhua},
  journal={arXiv preprint arXiv:1912.04486},
  year={2019}
}

@article{shi2023re,
  title={How re-sampling helps for long-tail learning?},
  author={Shi, Jiang-Xin and Wei, Tong and Xiang, Yuke and Li, Yu-Feng},
  journal={Advances in Neural Information Processing Systems},
  volume={36},
  pages={75669--75687},
  year={2023}
}

\clearpage
\appendix
\setcounter{section}{0}
\renewcommand{\thesection}{\Alph{section}}

\section{Technical Appendices and Supplementary Material}

\begin{figure}[!b]
\centering
% \begin{minipage}{0.6\columnwidth}
%     \subfloat[CCUA Framework]{
%       \label{cldm}
%         \includegraphics[width=0.95\columnwidth]{figures_iclr/cldm.pdf}}
% \end{minipage}
\begin{minipage}{1.0\columnwidth}
    \subfloat[w/ UNet-based Model]{
      \label{cldm_unet} \includegraphics[width=0.48\columnwidth]{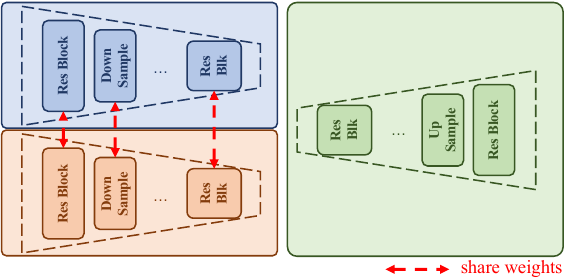}}
    \subfloat[w/ Diffusion Transformer]{
      \label{cldm_sit} \includegraphics[width=0.48\columnwidth]{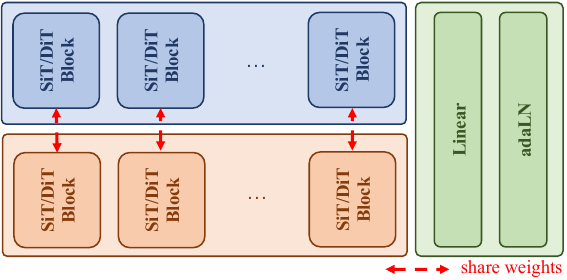}}
\end{minipage}

\caption{\fang{Model details of CCUA framework with UNet-based Model and Diffusion Transformer. As shown in Fig.~\ref{fig:method}, the noise estimation network is divided into two parts, a latent encoded network $e(*)$ and a decoded network $d(*)$.
(a) For UNet-based architecture, $e(*)$ is defined as the UNet encoder, while $d(*)$ is defined as the UNet decoder.
(b) For Diffusion Transformer, $e(*)$ is defined as all the SiT/DiT blocks, while $d(*)$ is defined as the final linear and adaLN projection layer.
}}
%\vspace{-6mm}
\label{fig:method_details}
\end{figure}

\begin{table*}[!b]
\caption{\ICLRrebuttal{Comparison on ImageNet-LT~$256 \times 256$ with SiT pipeline with imbalanced factor 0.001. CCUA results are based on CCUA-N models.}}
%\vspace{-3mm}
\label{tab: imb-factor 0.001}
\begin{center}
\setlength{\tabcolsep}{0.6mm}{
\small
\begin{tabular}{c|c|c|l|c|c|l}
\hline
\hline

Steps & Method & IS $\uparrow$ & FID $\downarrow$ & sFID $\downarrow$ & Prec. (\%) $\uparrow$ & Recall (\%) $\uparrow$ \\
\hline

\multirow{2}{*}{250k} & SiT & 29.89 & 51.26 & 21.26 & 38.09 & 22.04 \\
& \textbf{CCUA (ours)} & \textbf{48.79} & \textbf{36.41} \textcolor{blue}{(-14.85)} & \textbf{16.41} & \textbf{45.04} & 21.26 \\

\hline

\multirow{2}{*}{450k} & SiT & 46.47 & 37.55 & 23.50 & 48.66 & 20.39 \\
& \textbf{CCUA (ours)} & \textbf{83.93} & \textbf{25.96} \textcolor{blue}{(-11.59)} & \textbf{17.44} & \textbf{55.91} & 19.33 \\

\hline

\multirow{2}{*}{700k} & SiT & 59.05 & 30.56 & 22.05 & 54.79 & 20.61 \\
& \textbf{CCUA (ours)} & \textbf{105.86} & \textbf{22.44} \textcolor{blue}{(-8.12)} & \textbf{18.40} & \textbf{60.34} & 19.29 \\

\hline

\multirow{2}{*}{900k} & SiT & 67.70 & 27.25 & 19.96 & 57.49 & 18.89 \\
& \textbf{CCUA (ours)} & \textbf{118.16} & \textbf{20.93} \textcolor{blue}{(-6.32)} & \textbf{18.88} & \textbf{62.88} & 18.89 \\

\hline
\hline
\end{tabular}
}
%\vspace{-6mm}
\end{center}
\end{table*}

\begin{table}[!t]
\caption{Our method outperforms other baselines on all datasets. We also provide the results of DDPM trained on balanced datasets, which show the upper bound of performance. All models are measured with DDIM~\cite{ddim} 100 steps for conditional generation with CFG. Blue `\textcolor{blue}{()}' shows improvement of our method over DDPM baseline~\cite{ddpm}. Green `\textcolor{ForestGreen}{()}' shows improvement of DDPM trained on balanced version over trained on long-tailed version.}
\label{tab:comparison-lt-ddim}
%\vspace{-3mm}
\begin{center}
\setlength{\tabcolsep}{2mm}{

\small
\begin{tabular}{l|l|l|l|l}
\hline
\hline
Dataset & Method & FID$\downarrow$ & FID$_{tail}\downarrow$ & KID$_{\times1k}\downarrow$ \\%& Recall$_{tail}$ \\

\hline
\multirow{5}{*}{CIFAR10-LT}
& DDPM$^*_{bal}$~\cite{ddpm} & 4.90 \textcolor{ForestGreen}{(-1.03)} & 6.27 \textcolor{ForestGreen}{(-5.98)} & 1.32 \textcolor{ForestGreen}{(-0.32)} \\%& 0.559 \\
\cline{2-5}
& DDPM~\cite{ddpm} & 5.93 & 12.25 & 1.64 \\%& 0.493 \\
% & DDIM & 14.72 & 7.04 & 5.96 & -3.53 & 29.91 & - & 18.12 \\
& CBDM~\cite{cbdm} & 5.81 & \textbf{10.01} & 1.58 \\%& \textbf{0.528} \\
& OCLT~\cite{oclt} & 6.10 & 11.13 & 1.58 \\%& 0.491 \\
% \cline{2-7}
& \textbf{CCUA (ours)} & \textbf{5.56} \textcolor{blue}{(-0.37)} & 10.03 \textcolor{blue}{(-2.22)} & \textbf{1.27} \textcolor{blue}{(-0.37)} \\%& 0.478 \\ %mse1.0 nce0.1
\hline
\multirow{5}{*}{CIFAR100-LT}
& DDPM$^*_{bal}$~\cite{ddpm} & 5.15 \textcolor{ForestGreen}{(-1.80)} & 8.97 \textcolor{ForestGreen}{(-8.48)} & 1.05 \textcolor{ForestGreen}{(-0.66)} \\%& 0.560 \\
\cline{2-5}
& DDPM~\cite{ddpm} & 6.95 & 17.45 & 1.71 \\%& 0.436 \\
& CBDM~\cite{cbdm} & 6.50 & 17.36 & 1.41 \\%& 0.508 \\
& OCLT~\cite{oclt} & 6.45 & 17.22 & 1.42 \\%& 0.462 \\
% \cline{2-7}
& \textbf{CCUA (ours)} & \textbf{6.24} \textcolor{blue}{(-0.71)} & \textbf{16.35} \textcolor{blue}{(-1.10)} & \textbf{1.36} \textcolor{blue}{(-0.35)}\\ %& \textbf{0.509} \\%mse1.0 nce1.0

\hline
\hline
\end{tabular}
}
\end{center}
%\vspace{-6mm}
\end{table}

\subsection{Implementation Details}
\label{sec:details}

 \textbf{Dataset Details}\quad We keep the original $32\times32$ resolution for CIFAR10-LT and CIFAR100-LT, and resize images to $64 \times 64$ for TinyImageNet-LT and Places-LT, while $256 \times 256$ for ImageNet-LT.
 Same as~\cite{cbdm, diffrop, oclt}, we adopt the same methodology presented in~\cite{cifarlt} to construct long-tailed version datasets with a selected imbalance factor.
 Specifically, we conducted comparison experiments for imbalance factor $0.01$ (Tab.~\ref{tab: SiT on ImageNet-LT}) and $0.001$ (Tab.~\ref{tab: imb-factor 0.001}) on ImageNet-LT.

 \textbf{Batch Resample} is a simple strategy for long-tailed recognition and generation~\cite{oclt}. However, it may lead to mode collapse due to repetitive images for tail classes. Since our unsupervised contrastive loss relies on pairs of images, batch resampling increases the chance of images from the same tail class appearing in the same batch, which further diversify tail class images. We choose batch resampling as an optional strategy for our framework and provide an ablation study in Sec.~\ref{sec:visualization and analysis}. %Besides, we set batch re-sample technique as an optional strategy for InfoNCE loss, since a mini-batch with resampled, balanced samples would implicitly increase the weight of tail classes in $\mathcal{L}_{ucl}$, and further diverse the tail class samples.}

 \fang{\textbf{Model Architecture Details}\quad As described in Sec.~\ref{sec:method}, the proposed CCUA framework can be applied into UNet-based architecture and Diffusion Transformer. 
 In Fig.~\ref{fig:method_details}, we display our UNet-based model and Diffusion Transformer in details.
 As shown in Fig.~\ref{fig:method}, the noise estimation network is divided into two parts, a latent encoded network $e(*)$ and a decoded network $d(*)$.
 In our setting, for UNet-based model, $e(*)$ is defined as the UNet encoder, while $d(*)$ is defined as the UNet decoder, as shown in Fig.~\ref{fig:method_details} (a).
 For Diffusion Transformer, $e(*)$ is defined as all the SiT/DiT blocks, while $d(*)$ is defined as the final linear and adaLN projection layer, as shown in Fig.~\ref{fig:method_details} (b).}

 \textbf{Training Details}\quad For SiT pipeline, we strictly follow the same training hyper-parameter settings as Dispersive Loss~\cite{wang2025diffuse}.
 We use 4 RTX A6000ada GPUs to train all SiT based models on ImageNet-LT datasets with batch size $48$.
 During the training process, all methods are trained from scratch. We report 250k steps, 450k steps, 700k steps, and 900k steps results in Table~\ref{tab: SiT on ImageNet-LT}.
 For DDPM pipeline, we follow the same training configurations of the baseline models~\cite{ddpm, cbdm, oclt}. %, by setting diffusion scheduler $\beta_1=10^{-4}$ and $\beta_T=0.02$ with T=1000.
 % Adam~\cite{adam} optimizer is used to optimize the network while the learning rate is set to 0.0002 after 5000 iterations of the warmup process.
 We use one RTX 4090 GPU to train each model on CIFAR10-LT/CIFAR100-LT datasets with batch size $64$ while using 2 A100 GPUs for TinyImageNet-LT and Places-LT,  with batch size $128$.
 During the training process, all methods are trained from scratch for 200k iterations on CIFAR10-LT/CIFAR100-LT datasets, 100k iterations on TinyImageNet-LT and Places-LT datasets, while they follow the classifier-free guidance (CFG) algorithm~\cite{cfg}, which randomly drops labels with a probability of 10\%.
 On ImageNet-LT and TinyImageNet-LT datasets, we apply batch resample strategy with the re-balanced factor $0.1$, while we don't apply batch resample strategy on Places-LT and CIFAR10-LT/CIFAR100-LT datasets.
 \ICLRrebuttal{For all datasets}, we also apply timestep adaptive weight $t/T$ to unsupervised contrastive loss since we find the unsupervised contrastive loss could also benefit from such an adaptive weight, like alignment loss.

 \textbf{Evaluation Details}\quad In Table~\ref{tab: DDPM on TinyImageNet-LT and Places-LT}, FID$_{tail}$ denotes the FID score for synthetic images of the last 30\% classes in each dataset.
 Specifically, we categorize the `Tail' classes as the last 66 classes for TinyImageLT (200 classes), the last 121 classes for Places-LT (365 classes), respectively.
 % Specifically, we categorize the `Tail' classes as the last 3 classes for CIFAR10-LT, the last 33 classes for CIFAR100-LT, the last 66 classes for TinyImageLT (200 classes), the last 121 classes for Places-LT (365 classes), and the last 333 classes for ImageNetLT (1000 classes), respectively.
 %
 % For CIFAR10-LT/CIFAR100-LT datasets, the evaluation metrics are based on 50k synthetic images generated by each method, while for the ImageNetLT dataset, the evaluation metrics are based on 20k synthetic images generated by each method.
 In Table~\ref{tab: SiT on ImageNet-LT}, the evaluation metrics are based on 50k synthetic images generated by each method with a CFG strength 7.5.
 In Table~\ref{tab: DDPM on TinyImageNet-LT and Places-LT}, the evaluation metrics are based on 10k synthetic images generated by each method.
 During inference, we perform a grid search algorithm for each method to determine the optimal guidance strength $\omega$ of CFG, ensuring each model achieves its best performance.

 \subsection{\ICLRrebuttal{Accelerated Implementation}}
\label{sec:acceleratedimplementation}

 \ICLRrebuttal{CCUA incurs longer training time for diffusion models. We further optimized our implementation by a simple trick of computing conditional generation and unconditional generation for the same batch with one function call of model.forward(). In a naive implementation, we called model.forward() twice in each iteration, which is not as efficient. }

\begin{lstlisting}[language=Python]

# CCUA (serial)
# x_batch.shape: B, C, H, W
# original implementation for one iteration
cond_output = model.forward(x_batch, c)
uncond_output = model.forward(x_batch, null)

# CCUA (parallel)
# optimized implementation for one iteration
cond_output, uncond_output = model.forward(
    torch.cat([x_batch, x_batch], dim=0),
    torch.cat([c, null], dim=0)
).chunk(2, dim=0)
\end{lstlisting}

\ICLRrebuttal{At the cost of 1.18x GPU memory consumption (0.44 GB to 0.52GB), our optimized implementation is only 1.3x slower than SiT baseline with original diffusion loss.}

\begin{table}[!b]
\centering
\begin{tabular}{lcc}
\hline
\ICLRrebuttal{Method} & \ICLRrebuttal{Average Training Time} & \ICLRrebuttal{GPU Consumption} \\
\hline
\ICLRrebuttal{SiT} & \ICLRrebuttal{8.7 steps/s} & \ICLRrebuttal{0.44 GB/image} \\
\ICLRrebuttal{CCUA (serial)} & \ICLRrebuttal{5.8 steps/s} & \ICLRrebuttal{0.48 GB/image} \\
\ICLRrebuttal{CCUA (parallel)} & \ICLRrebuttal{6.7 steps/s} & \ICLRrebuttal{0.52 GB/image} \\
\hline
\end{tabular}
\end{table}

\begin{table}[!t]
\caption{We compare our method with other baselines on CIFAR10-LT/CIFAR100-LT datasets, with DDPM 1000 steps for conditional generation with CFG. We also report the data augmentation method ADA~\cite{karras2020training} and $\omega$-scheduler~\cite{omega_scheduler}, which are orthogonal to our method.}
%\vspace{-3mm}
\label{tab:comparison-lt-ddpm}
\begin{center}
\setlength{\tabcolsep}{6mm}{

\small
\begin{tabular}{l|c|c|c|c}
\hline
\hline
\multirow{2}{*}{Method} & \multicolumn{2}{c|}{CIFAR10-LT} & \multicolumn{2}{c}{CIFAR100-LT} \\
\cline{2-5}
& FID$\downarrow$ & IS$\uparrow$ & FID$\downarrow$ & IS$\uparrow$ \\
\hline
 DDPM$^*_{bal}$~\cite{ddpm} & 4.87 & 9.35 & 5.20 & 13.29 \\
 \hline
 DDPM~\cite{ddpm} & 5.81 & 9.36 & 7.09 & 12.64 \\
 \quad+ADA~\cite{karras2020training} & - & - & 6.69 & 12.87 \\
 \quad+$\omega$-Scheduler~\cite{omega_scheduler} & 5.87 & 9.22 & 6.60 & 12.10 \\
 
 CBDM~\cite{cbdm} & 5.92 & 9.38 & 6.52 & 12.79 \\
 OCLT~\cite{oclt} & 5.69 & 9.42 & 6.23 & \textbf{13.18} \\
\hline
 \textbf{CCUA (ours)} & \textbf{5.57} & \textbf{9.42} & \textbf{5.99} & 13.01 \\ %mse0.1 nce0.1
 % \quad+ADA & - & - & - & - & - & - \\
\hline
\hline
\end{tabular}
}
\end{center}
\end{table}

\subsection{More Quantitative Results}
\label{sec:quantative appendix}

 \ICLRrebuttal{\textbf{Extremely Imbalanced Generation on ImageNet-LT dataset}\quad We measure the performance of CCUA and SiT w.r.t the imbalanced factor 0.001 for ImageNet-LT dataset. Note that for ImageNet datasets, each class only contains 1300 images, which means with 0.001 imbalanced factor the tail classes only contains 1$\sim$2 images. Our method improves the baseline for such a challenging dataset.}

 \textbf{Class-imbalanced Generation on CIFAR-LT datasets}\quad We conduct more experiments and analysis on CIFAR10-LT/CIFAR100-LT datasets, as shown in Table~\ref{tab:comparison-lt-ddim} and Table~\ref{tab:comparison-lt-ddpm}.
 We provide the metrics of the DDPM model trained on the balanced version, i.e., the original CIFAR10/CIFAR100 datasets, denoted by DDPM$^*_{bal}$, as the theoretical optimal reference.
 On CIFAR10-LT/CIFAR100-LT, our method achieves the lowest FID and KID compared to baseline methods.
 Note that the FID gap between DDPM and DDPM$^*_{bal}$ is \textbf{1.03} on CIFAR10-LT and \textbf{1.8} on CIFAR100-LT, while our method improves FID \textbf{0.37} over 1.03 on CIFAR10-LT and \textbf{0.71} over 1.8 on CIFAR100-LT, achieving $>\textbf{35\%}$ performance improvement.
 To investigate the consistency of such improvements, we compare our method and all baseline methods on CIFAR10-LT and CIFAR100-LT with DDPM 1000 steps. As shown in Tab.~\ref{tab:comparison-lt-ddpm}, our method achieves consistent improvements of FID for full 1000 sampling steps.
 We also compare to a widely used data augmentation technique, Adaptive Discriminator Adaption (ADA)~\cite{karras2020training} for generative models on the full DDPM 1000 sampling steps.
 Besides, we apply the CFG guidance strength scheduler $\omega$-cos~\cite{omega_scheduler} on DDPM, which gradually increases the guidance strength during sampling time steps decreasing to force the model transfer from unconditional to conditional generation.
 % Even though the proposed method is orthogonal to the data augmentation methods, we believe the comparison between our method and the methods could lead to new insights on how the diffusion-based generative models could benefit from knowledge transfer from unconditional to conditional models.
 % }}
% \TODO{[Move to Appendix.] For CIFAR10-LT/CIFAR100-LT and TinyImageNet-LT datasets, we provide the metrics of the DDPM model trained on the balanced version, i.e., CIFAR10/CIFAR100 and TinyImageNet datasets, denoted by DDPM$^*_{bal}$, as a lower bound reference.}
 %

\textbf{Consistent Improvement for Fewer Sampling Steps}\quad
 To investigate the model's performance on extremely few sampling steps, We evaluate our method and all baselines for DDIM 10 steps with CFG conditional generation on CIFAR10-LT/CIFAR100-LT.
 As shown in Tab.~\ref{tab:comparison-lt-ddim-10}, our method achieves the best FID, FID$_{tail}$ and KID scores.
 Our method achieves even better results than the theoretical optimal model DDPM$^*_{bal}$ under such an extreme experimental setting.
 Such an improvement demonstrates the effectiveness of our method for training long-tailed image generation diffusion model.

%\vspace{-3mm}
\begin{table}[!t]
\caption{We compare our proposed $\mathcal{L}_{ccua}$ loss with other baselines on CIFAR10-LT/CIFAR100-LT datasets with DDIM 10 sampling steps for conditional generation with CFG. Blue `\textcolor{blue}{()}' shows improvement of our method over DDPM baseline~\cite{ddpm}. Green `\textcolor{ForestGreen}{()}' shows improvement of DDPM trained on balanced version over trained on long-tailed version.}
%\vspace{-3mm}
\label{tab:comparison-lt-ddim-10}
\begin{center}
\setlength{\tabcolsep}{2mm}{

\small
\begin{tabular}{l|l|l|l|l}
\hline
\hline
Dataset & Method & FID$\downarrow$ & FID$_{tail}\downarrow$ & KID$_{\times1k}\downarrow$ \\
\hline
\multirow{5}{*}{CIFAR10-LT} 
& DDPM$^*_{bal}$ & 13.28~\textcolor{ForestGreen}{(-1.44)} & 13.26~\textcolor{ForestGreen}{(-5.01)} & 6.06~\textcolor{ForestGreen}{(-0.98)} \\
\cline{2-5}
& DDPM & 14.72 & 18.27 & 7.04 \\
% & DDIM & 14.72 & 7.04 & 5.96 & -3.53 & 29.91 & - & 18.12 \\
& CBDM & 13.54 & 16.90 & 6.52 \\
& OCLT & 15.48 & 20.73 & 7.39 \\
% \cline{2-7}
& \textbf{CCUA (ours)} & \textbf{13.16}~\textcolor{blue}{(-1.56)} & \textbf{16.83}~\textcolor{blue}{(-1.44)} & \textbf{6.04}~\textcolor{blue}{(-1.00)} \\ %mse1.0 nce0.1
\hline
\multirow{5}{*}{CIFAR100-LT} 
& DDPM$^*_{bal}$ & 13.34~\textcolor{ForestGreen}{(-0.75)} & 18.27~\textcolor{ForestGreen}{(-6.80)} & 5.56~\textcolor{ForestGreen}{(-0.57)} \\
\cline{2-5}
& DDPM & 14.09 & 25.07 & 6.13 \\
& CBDM & 13.37 & 23.97 & 5.83 \\
& OCLT & 13.70 & 24.48 & 5.73 \\
% \cline{2-7}
& \textbf{CCUA (ours)} & \textbf{12.90}~\textcolor{blue}{(-1.19)} & \textbf{23.17}~\textcolor{blue}{(-1.90)} & \textbf{5.63}~\textcolor{blue}{(-0.50)} \\ %mse1.0 nce1.0
\hline
\hline
\end{tabular}
}
%\vspace{-6mm}
\end{center}
\end{table}

\setlength{\intextsep}{-12pt}
\begin{wraptable}{r}{6.0cm}
%\vspace{-12mm}
\caption{Unconditional generation w/ DDIM 100 steps.}
\label{tab:comparison-lt-uncond}
%\vspace{-2mm}
\begin{center}
\setlength{\tabcolsep}{1.6mm}{

\small 
\begin{tabular}{l|c|c|c|c}
\hline
\hline
\multirow{2}{*}{Method} & \multicolumn{2}{c|}{CIFAR10-LT} & \multicolumn{2}{c}{CIFAR100-LT} \\ 
\cline{2-5}
& FID$\downarrow$ & IS$\uparrow$ & FID$\downarrow$ & IS$\uparrow$ \\
\hline
 DDPM & 27.52 & 6.65 & 18.53 & 8.68 \\
 CBDM & 25.60 & 6.70 & 17.06 & 9.00 \\
 OCLT & 31.38 & 6.34 & 18.97 & 8.73 \\
\hline
\textbf{Ours} & \textbf{24.16} & \textbf{6.80} & \textbf{15.97} & \textbf{9.23} \\ %mse0.1 nce0.1
\hline
\hline
\end{tabular}
}
%\vspace{-10mm}
\end{center}
\end{wraptable}

\textbf{Class-imbalanced Unconditional Generation}\quad
 We also evaluate all the models for class-imbalanced unconditional generation.
 As shown in Tab.~\ref{tab:comparison-lt-uncond}, the proposed method reduces FID from \textbf{27.52} to \textbf{24.16} for CIFAR10-LT and from \textbf{18.53} to \textbf{15.97} for CIFAR100-LT.
 Such an improvement in unconditional generation highlights the effectiveness of the proposed contrastive learning loss, particularly the unsupervised contrastive loss $\mathcal{L}_{ucl}$.

\textbf{Comparison to Other Baseline Methods} In Table~\ref{tab: diffrop}, we provide comparision to DiffROP~\cite{diffrop}, which is another method for regularizing diffusion model with long-tailed training data.

\begin{table*}[!t]
\caption{\ICLRrebuttal{Comparison with DiffROP~\cite{diffrop} on ImageNet-LT.}}
%\vspace{-3mm}
\label{tab: diffrop}
\begin{center}
\setlength{\tabcolsep}{0.6mm}{
\small
\begin{tabular}{c|c|c|c|c|c|c}
\hline
\hline

Steps & Method & IS $\uparrow$ & FID $\downarrow$ & sFID $\downarrow$ & Prec. (\%) $\uparrow$ & Recall (\%) $\uparrow$ \\
\hline

\multirow{2}{*}{250k} & SiT & 53.9 & 33.8 & 22.6 & 54.5 & 19.1 \\
& DiffROP & 53.8 & 34.3 & 24.1 & 53.8 & 19.6 \\
& \textbf{CCUA (ours)} & \textbf{70.7} & \textbf{27.0} & \textbf{20.5} & \textbf{60.2} & \textbf{20.1} \\

\hline

\multirow{2}{*}{450k} & SiT & 78.8 & 25.7 & 21.9 & 64.3 & 18.5 \\
& DiffROP & 76.4 & 28.1 & 29.1 & 62.4 & 18.1 \\
& \textbf{CCUA (ours)} & \textbf{111.9} & \textbf{19.4} & \textbf{18.3} & \textbf{69.4} & \textbf{18.9} \\

\hline

\multirow{2}{*}{700k} & SiT & 103.1 & 21.2 & 20.1 & 69.3 & 18.2 \\
& DiffROP & 97.4 & 23.3 & 26.9 & 67.6 & 18.1 \\
& \textbf{CCUA (ours)} & \textbf{140.5} & \textbf{16.3} & \textbf{17.2} & \textbf{73.9} & \textbf{18.4} \\

\hline

\multirow{2}{*}{900k} & SiT & 111.7 & 19.9 & 20.1 & 70.3 & 18.6 \\
& DiffROP & 107.7 & 21.6 & 25.6 & 70.0 & 17.0 \\
& \textbf{ CCUA (ours) } & \textbf{153.1} & \textbf{15.1} & \textbf{16.5} & \textbf{75.7} & \textbf{17.3} \\

\hline
\hline
\end{tabular}
}
%\vspace{-6mm}
\end{center}
\end{table*}

\subsection{More Ablation Study}
\label{sec:ablation_study_appendix}

\begin{table*}[!t]
\caption{Batch Resample Strategy (BRS) Analysis on ImageNet-LT. Blue `\textcolor{blue}{()}' highlights the improvement of each method compared to the SiT baseline, while Red `\textcolor{red}{()}' highlights the decline v.s. SiT baseline.}
\label{tab: batch resample strategy analysis}
\begin{center}
\setlength{\tabcolsep}{0.5mm}{
\small
\begin{tabular}{l|l|c|c|c|c|c}
\hline
\hline

 Steps & Method & IS $\uparrow$ & FID $\downarrow$ & sFID $\downarrow$ & Prec. (\%) $\uparrow$ & Recall (\%) $\uparrow$ \\
\hline

\multirow{5}{*}{250k} & SiT (baseline) & 53.9 & 33.8 & 22.6 & 54.5 & 19.1 \\
& \quad + BRS & 71.4 & 28.1 \textcolor{blue}{(-5.7)} & 21.6 \textcolor{blue}{(-1.0)} & 57.6 & 17.3 \textcolor{red}{(-1.8)} \\
\cline{2-7}
& $\mathcal{L}_{ccua}$ (Eq.~\ref{eq:final loss}) & 57.8 & 32.2 \textcolor{blue}{(-1.6)} & 20.7 \textcolor{blue}{(-1.9)} & 54.5 & 20.7 \textcolor{blue}{(+1.6)} \\
& $\mathcal{L}_{ccua(N/4)}$ + BRS & 70.7 & 27.0 \textcolor{blue}{(-6.8)} & 20.5 \textcolor{blue}{(-2.1)} & \textbf{60.2} & 20.1 \textcolor{blue}{(+1.0)} \\
& $\mathcal{L}_{ccua(N)}$ + BRS & \textbf{73.6} & \textbf{25.9} \textcolor{blue}{(-7.9)} & \textbf{16.5} \textcolor{blue}{(-6.1)} & 58.6 & \textbf{23.1} \textcolor{blue}{(+4.0)} \\

\hline

\multirow{3}{*}{900k} & SiT (baseline) & 111.7 & 19.9 & 20.1  & 70.3 & \textbf{18.6} \\
& + BRS & 139.6 & 16.7& 20.8 & 73.6 & 16.6 \\
% & $\mathcal{L}_{ccua}$ (Eq.~\ref{eq:final loss}) &  \\
& + BRS + $\mathcal{L}_{ccua}$ & \textbf{153.1} & \textbf{15.1} & \textbf{16.5} & \textbf{75.7} & 17.3\\

\hline
\hline
\end{tabular}
}
\end{center}
\end{table*}

 \textbf{Ablation Study of Batch Resample Strategy on ImageNet-LT}\quad We conduct ablation study of batch resample strategy on ImageNet-LT dataset, as shown in Table~\ref{tab: batch resample strategy analysis}, where the proposed $\mathcal{L}_{ccua}$ consistently outperforms SiT with and without applying batch resampling strategy, separately. We also compare the performance of $\mathcal{L}_{ccua}$ used on latents from different DiT/SiT block ($N/4$-th, $N$-th) with applying batch resampling strategy.

 \textbf{Discussion of Latent Encoder of SiT}\quad We conduct the ablation study to investigate the choice of latent encode network in SiT. 
 For a SiT model with $N$ SiT/DiT blocks, we use the latent representations $h$ from the $N/4$-th, $N/2$-th, $3N/4$-th, and the $N$-th block for $\mathcal{L}_{ucl}$ loss. 
 As shown in Table~\ref{tab: latent encoder}, we found even the N/4-th layer does not always provide the best performance on all metrics. N-th layer, i.e., the last layer, generally has better spatial FID (sFID) score and Recall but worse overall FID compared to N/4-th layer.
 These two metrics reflect the latent space quality and diversity of the generated images.
 In order to trade off the overall FID with sFID/Recall, we select the N/4-th layer as the mainly reported encoder.

\begin{table*}
\caption{\ICLRrebuttal{Ablation Study of Latent Encoder for DiT-based model.}}
%\vspace{-3mm}
\label{tab: latent encoder}
\begin{center}
\setlength{\tabcolsep}{2mm}{
\small
\begin{tabular}{c|c|c|c|c|c|c}
\hline
\hline

Steps & Latent & IS $\uparrow$ & FID $\downarrow$ & sFID $\downarrow$ & Prec. (\%) $\uparrow$ & Recall (\%) $\uparrow$ \\
\hline

\multirow{4}{*}{250k} 
& \ICLRrebuttal{\textbf{N/4}} & 70.69 & 27.04 & 20.50 & \textbf{60.16} & 20.12 \\
& \ICLRrebuttal{\textbf{N/2}} & 68.57 & 28.00 & 21.13 & 58.59 & 21.28 \\
& \ICLRrebuttal{\textbf{3N/4}} & 68.48 & 27.99 & 19.96 & 58.96 & 21.09 \\
& \textbf{ N } & \textbf{73.60} & \textbf{25.89} & \textbf{16.54} & 58.62 & \textbf{23.05}  \\

\hline

\multirow{4}{*}{450k} 
& \ICLRrebuttal{\textbf{N/4}} & \textbf{111.91} & \textbf{19.43} & 18.25 & \textbf{69.44} & 18.88 \\
& \ICLRrebuttal{\textbf{N/2}} & 105.41 & 20.10 & 19.49 & 68.22 & 19.24 \\
& \ICLRrebuttal{\textbf{3N/4}} & 109.29 & 19.78 & 18.50 & 69.30 & 18.80 \\
& \textbf{ N } & 103.66 & 19.99 & \textbf{14.88} & 65.80 & \textbf{21.31}  \\

\hline

\multirow{4}{*}{700k} 
& \ICLRrebuttal{\textbf{ N/4 }} & \textbf{140.46} & \textbf{16.29} & 17.19 & \textbf{73.92} & 18.40 \\
& \ICLRrebuttal{\textbf{ N/2 }} & 134.89 & 16.63 & 17.89 & 72.94 & 18.86 \\
& \ICLRrebuttal{\textbf{ 3N/4 }} & 137.21 & 16.65 & 17.46 & 73.60 & 18.41 \\
& \textbf{ N } & 119.11 & 17.55 & \textbf{13.89} & 68.38 & \textbf{21.20}  \\

\hline

\multirow{4}{*}{900k} 
& \ICLRrebuttal{\textbf{ N/4 }} & \textbf{153.07} & \textbf{15.08} & 16.54 & \textbf{75.73} & 17.27 \\
& \ICLRrebuttal{\textbf{ N/2 }} & 148.61 & 15.14 & 16.94 & 75.40 & 18.11 \\
& \ICLRrebuttal{\textbf{ 3N/4 }} & 148.88 & 15.22 & 16.80 & 75.32 & 19.35 \\
& \textbf{ N } & 124.87 & 16.41 & \textbf{13.17} & 69.84 & \textbf{21.34} \\

\hline
\hline
\end{tabular}
}
%\vspace{-6mm}
\end{center}
\end{table*}

 \textbf{Discussion of Time Scheduler.}\quad We conduct the ablation study to investigate the choice of different time-adaptive weight of alignment loss. As shown in Table~\ref{tab: time scheduler}, linear scheduler achieves the best performance.

\begin{table}[!tb]
\caption{Time Scheduler for the Weight of Alignment Loss. Results are reported from 900k training steps.}
\label{tab: time scheduler}
\begin{center}
\setlength{\tabcolsep}{2mm}{
\small
\begin{tabular}{l|c|c|c|c|c}
\hline
\hline

Scheduler & IS $\uparrow$ & FID $\downarrow$ & sFID $\downarrow$ & Prec. \% $\uparrow$ & Recall \% $\uparrow$ \\
\hline

Constant  & 147.5 & 15.7 & 20.8 & 73.8 & \textbf{18.3} \\
Quadratic  & 145.4 & 16.1 & 21.0 & 74.8 & 17.1 \\
\textbf{ Linear } & \textbf{153.1} & \textbf{15.1} & \textbf{16.5} & \textbf{75.7} & 17.3 \\

\hline
\hline
\end{tabular}
}
\end{center}
\end{table}

\newpage
\subsection{More Qualitative Results}
\label{sec:qualitative appendix}

\subsubsection{More Visualization of Synthetic Images on ImageNet-LT and Tiny ImageNet-LT}
\label{sec:more visualization} 
\begin{figure}
\centering
\begin{minipage}{0.98\columnwidth}
    \subfloat[Synthetic images from SiT~\cite{ma2024sit}.]{
        \includegraphics[width=0.98\columnwidth]{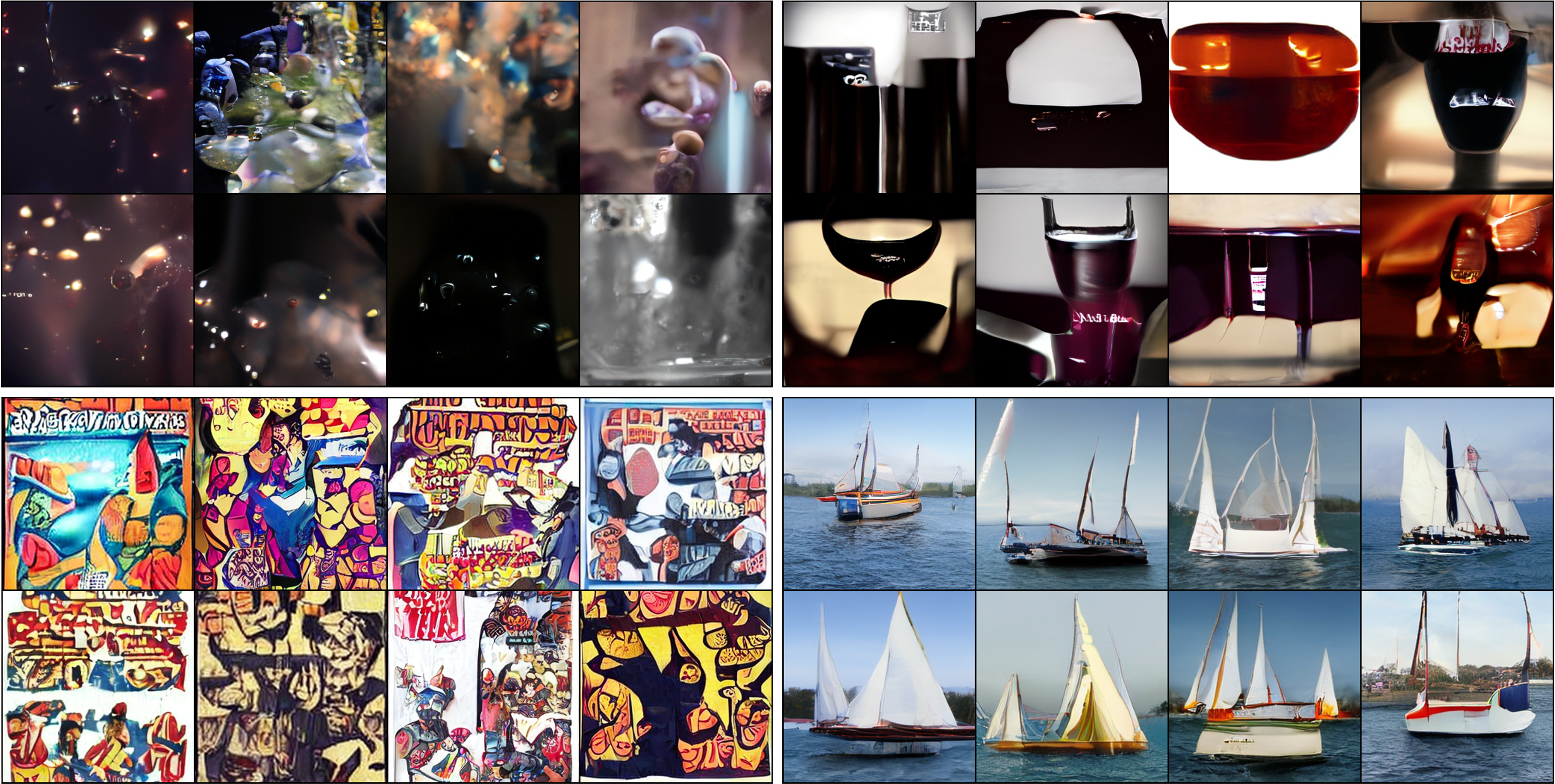}
      }
\end{minipage}
\begin{minipage}{0.98\columnwidth}
    \subfloat[Synthetic images from CCUA (ours).]{
        \includegraphics[width=0.98\columnwidth]{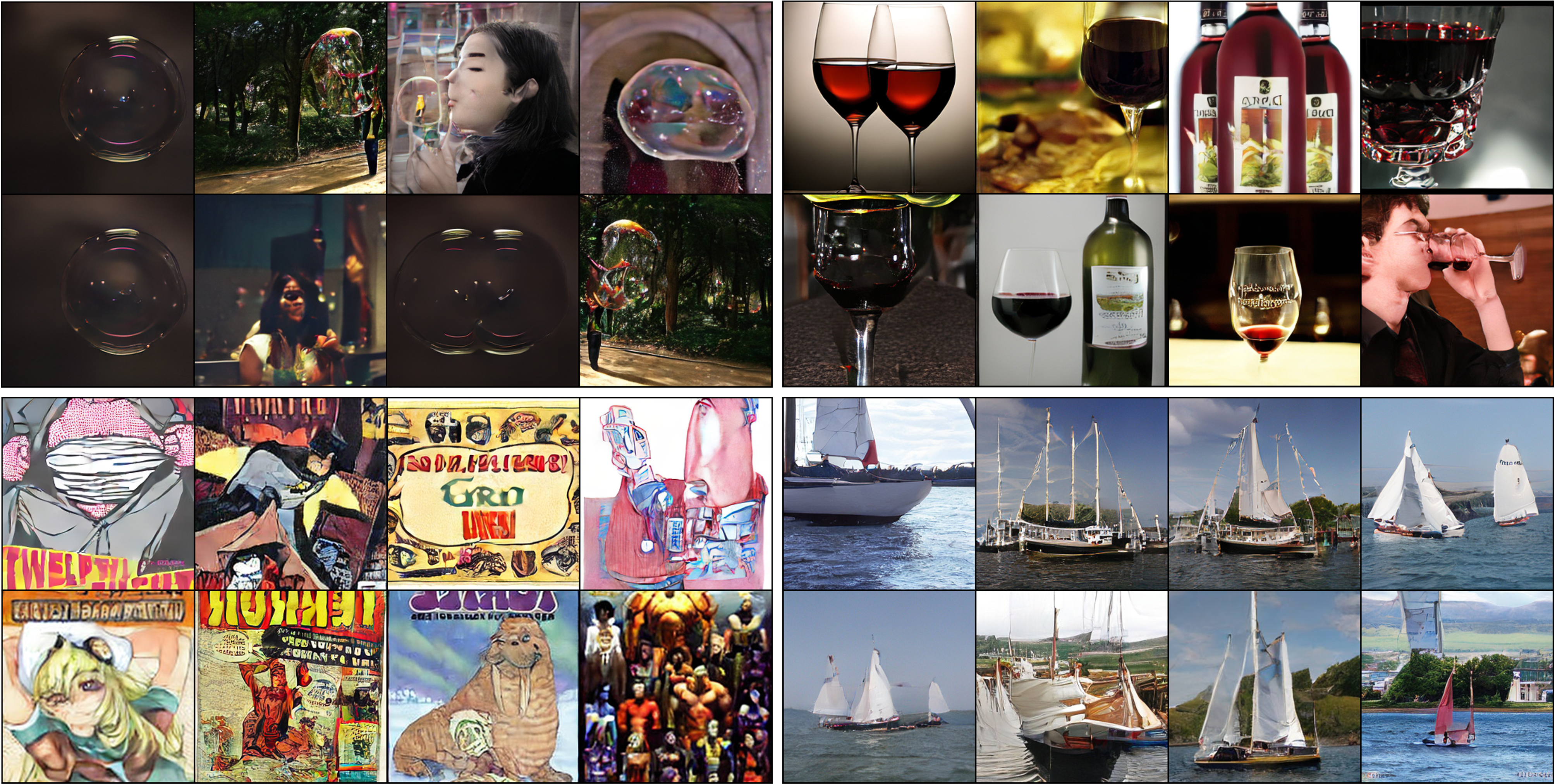}
      }
\end{minipage}

\caption{Synthetic images of SiT and CC UA for ImageNet-LT tail classes (from top-left to right-bottom: `bubble', `redwine', `comic book' and `yawl').
All methods start the denoising process from the same Gaussian noise at corresponding grid cells.
CCUA shows consistently higher diversity and fidelity compared to SiT.}
\label{fig:synthetic images on imagenet}
\end{figure}

\begin{figure*}[thbp]
    \centering
    \begin{minipage}{0.9\columnwidth}
        \subfloat[Synthetic images from DDPM~\cite{ddpm}.]{
            \includegraphics[trim=0 0 0 0, clip, width=1.0\textwidth]{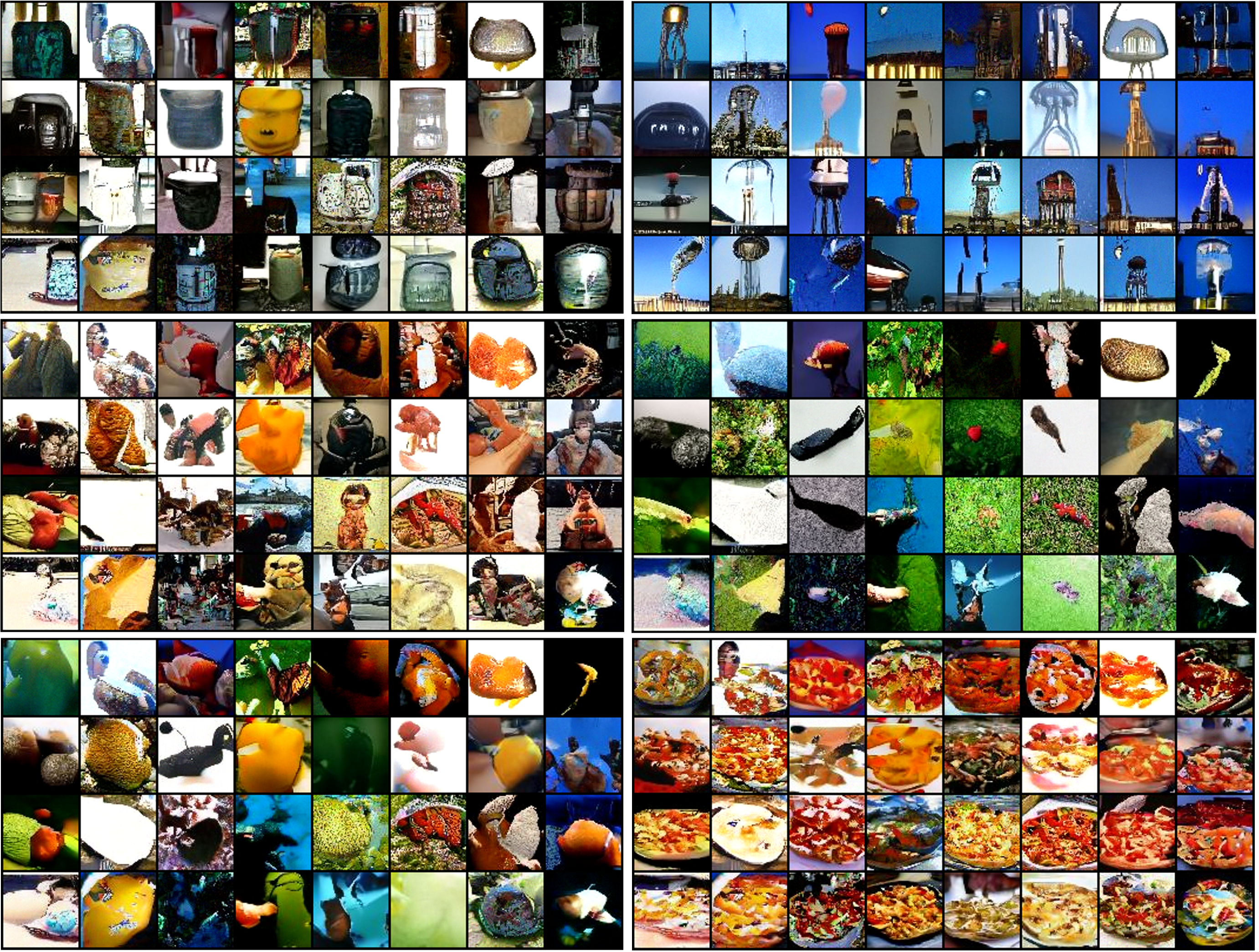}
        }
    \end{minipage}
    
    \begin{minipage}{0.9\columnwidth}
        \subfloat[Synthetic images from CCUA (ours).]{
        \includegraphics[trim=0 0 0 0, clip, width=1.0\textwidth]{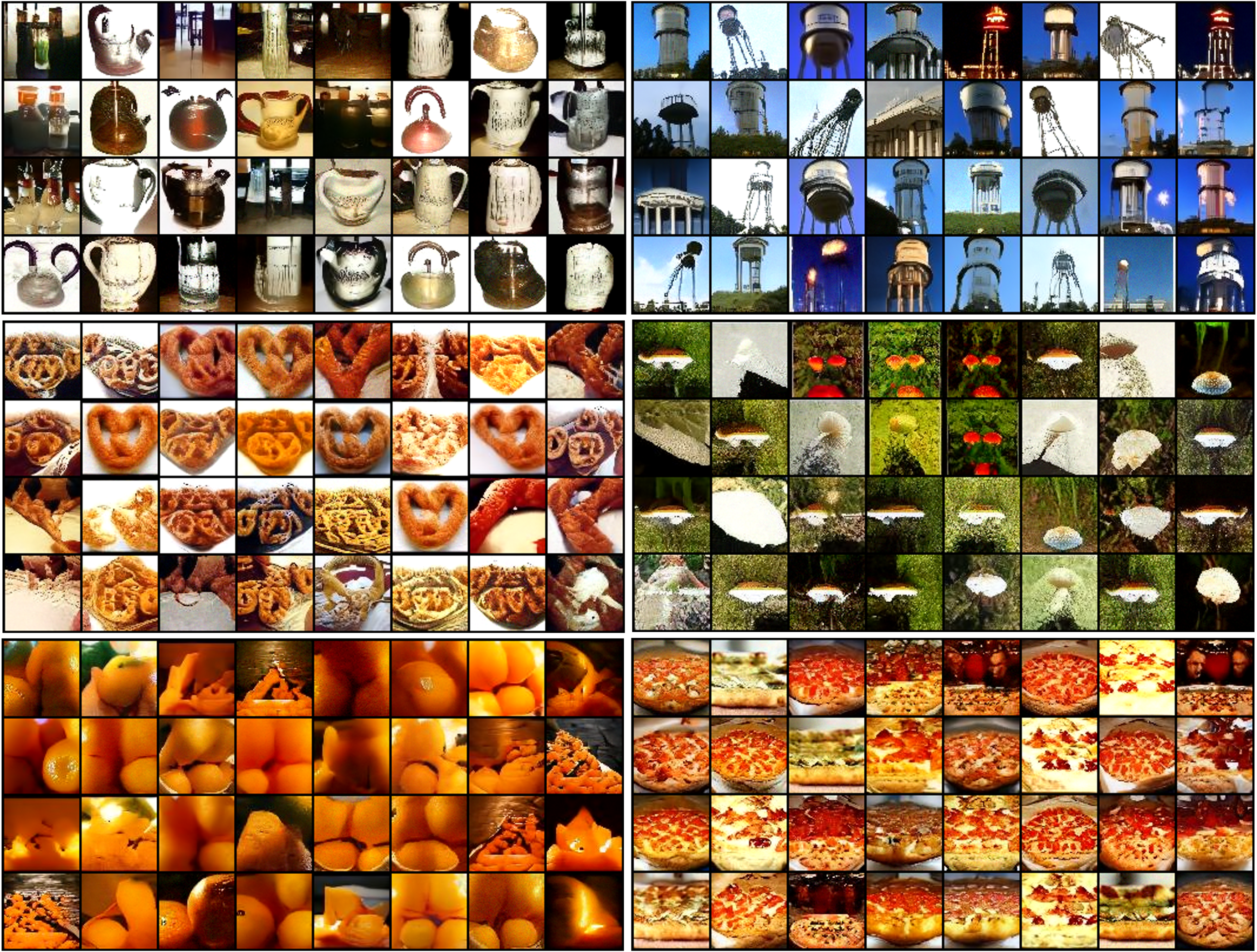}
    }
    \end{minipage}

    \caption{
    More synthetic results for TinyImageNet-LT tail classes (from top-left to right-bottom: `teapot', `water tower', `pretzel', `mushroom', `orange', and `pizza').
    Images in corresponding grid cell for DDPM and our method are initialized from the same Gaussian noise. Our method achieves more diverse images with higher fidelity for tail classes. 
    \fang{Note that for `pretzel' and `orange' classes, DDPM fails to synthesize images correlated to the class while CCUA synthesizes diverse images with high quality.}
    }
    \label{fig:synthetic images on tinyimagenet}
\end{figure*}

 Fig.~\ref{fig:synthetic images on imagenet} shows synthetic images of SiT and with the proposed CCUA for ImageNet-LT tail classes `bubble', `redwine', `comic book' and `yawl'.
 Fig.~\ref{fig:synthetic images on tinyimagenet} shows synthetic images of DDPM and with the proposed CCUA for TinyImageNet-LT tail classes `teapot', `water tower', `pretzel', `mushroom', `orange', and `pizza'.
 These methods start the denoising process from the same Gaussian noise at corresponding grid cells.
 As shown in Fig.~\ref{fig:synthetic images on imagenet} and Fig.~\ref{fig:synthetic images on tinyimagenet}, synthetic images of CCUA show consistently higher diversity and fidelity compared to SiT.

\begin{figure}[!t]
% \begin{wrapfigure}{r}{7.5cm}
% %\vspace{-7mm}
    \centering

    \begin{minipage}{0.99\columnwidth}
    \subfloat[Images in low-dimensional embeddings.]{
    \includegraphics[height=3.16cm]{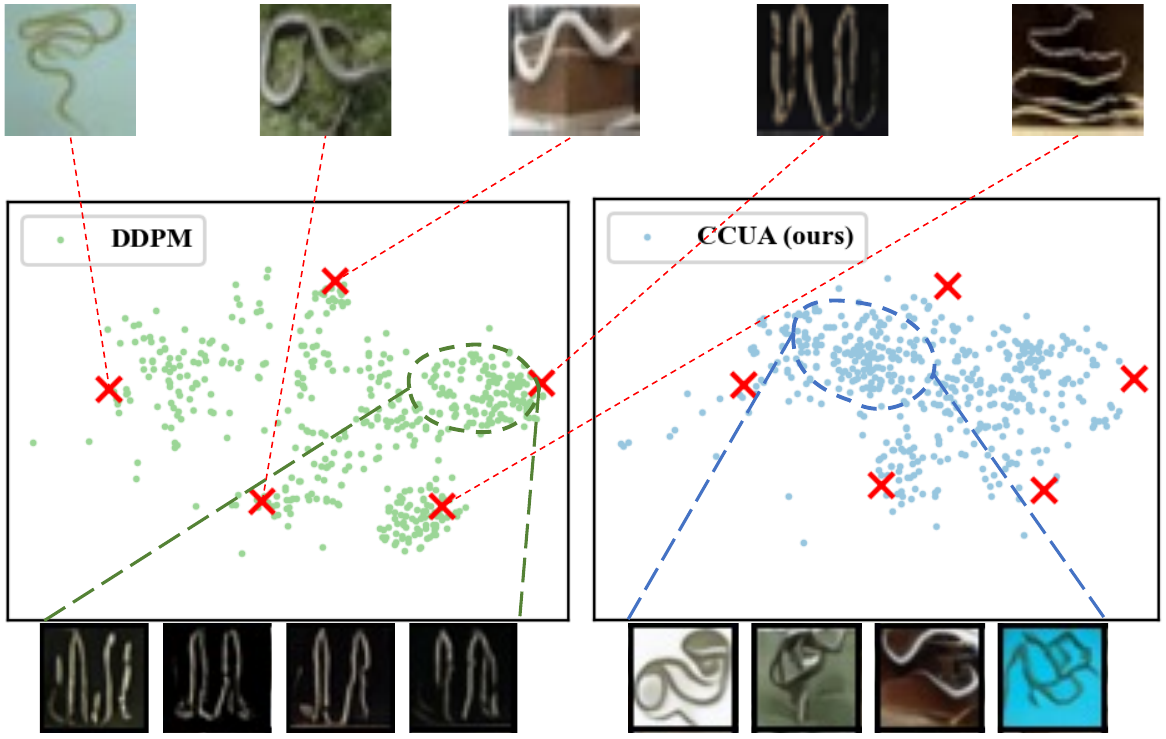}}
    \subfloat[Density distribution of synthetic images.]{
    \includegraphics[height=2.56cm]{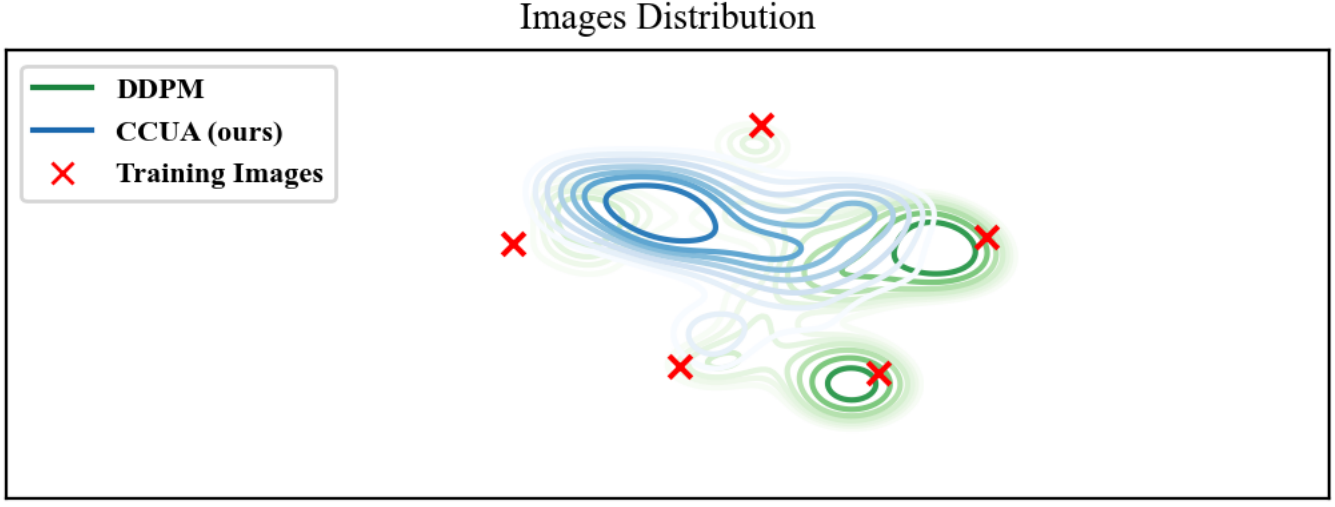}
    }
    \end{minipage}
    \caption{
    (a) Visualization of low-dimensional embeddings of five training images \textcolor{red}{x} for a tail class (`worm' in CIFAR100-LT) and synthetic images generated by DDPM~\cite{ddpm} and our method. 
    % The original DDPM overfits and generates highly similar images (top-left). 
    % In contrast, synthetic images based on our method have more diversity (top-right).
    (b) We also show the distributions of real images and synthetic images generated by our method and the original DDPM.
    % Greater color saturation (greener or bluer) indicates regions of higher density or modes within the image distribution.
    % The grey region denotes the real data distribution of `worm' images in the class-balanced CIFAR100 dataset.
    % Compared to DDPM, the distribution of synthetic images from our model is closer to the real distribution of images from the balanced dataset.
    }
    \label{fig:distribution}
\end{figure}
% %\vspace{-7mm}
% \end{wrapfigure}

% \newpage
\subsubsection{Distribution of Synthetic Images for Tail Class}
\label{sec:distribution}
 We visualize synthetic images in the feature space and their density function for our method and the original DDPM.
 Specifically, we use the Inception-V3~\cite{szegedy2016rethinking} model to extract $2048$ dimensional features of each image, and then apply t-SNE~\cite{tsne} to project these features into $2$ dimensions.
 As shown in Fig.~\ref{fig:distribution}, the original DDPM overfits and generates highly similar images, while synthetic images based on our method have more diversity.
 We visualize the image distribution more specifically by using kernel density estimation in the bottom.
 The grey region represents the distribution of real `worm' images from the class-balanced CIFAR100 dataset, while red `x' points denote all 5 `worm' images from the class-imbalanced CIFAR100-LT dataset.
 The blue region represents the distribution of images synthesized by our method, while the green region is for the original DDPM.
Areas with higher color saturation (dark, green, or blue) indicate regions of higher density, which correspond to modes of distribution.
 Synthetic images from DDPM mostly concentrates around the training images, leading to mode collapse.
 Synthetic images from our method spans the space enclosed by all training images, the distribution of which is shown in blue in (b).
 %Compared to DDPM, the image distribution of our method is closer to the real data distribution, which illustrates better distribution matching with balanced data.
 % By applying the proposed contrastive learning loss $\mathcal{L}_{ucl}$ to force the synthesized image instances far away from each other, and $\mathcal{L}_{al}$ to share rich features of head classes to tail classes, our method's synthesized images are sparser and uniformly distributed in the feature space with higher diversity, as shown in the bottom-left subfigure.

\subsubsection{Mode Collapse Issue on Tail Class}
\label{sec:mode collapse issue}

\begin{figure*}[b!]
    \centering

    \includegraphics[trim=0 0 0 0, clip, width=1.0\textwidth]{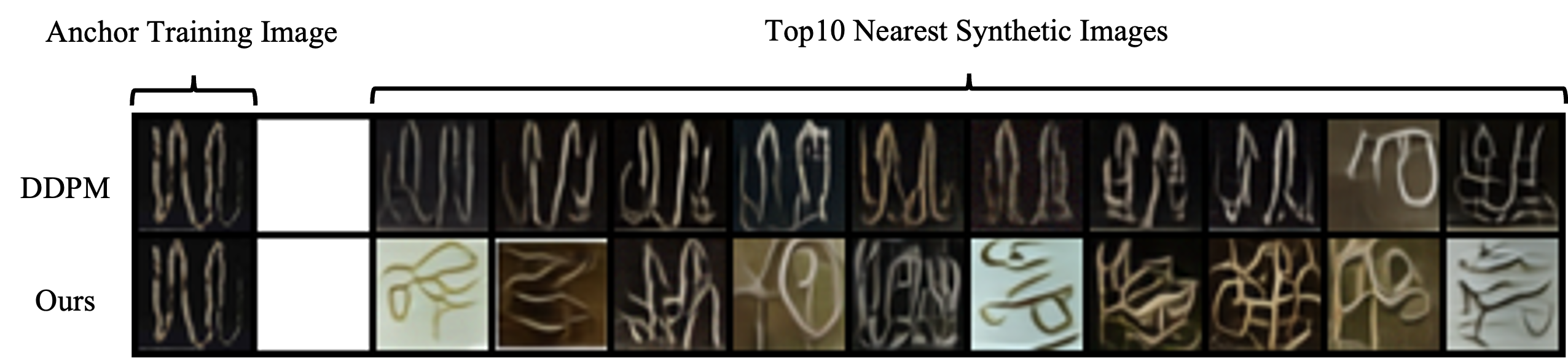}\\[1ex]

    \includegraphics[trim=0 0 0 0, clip, width=1.0\textwidth]{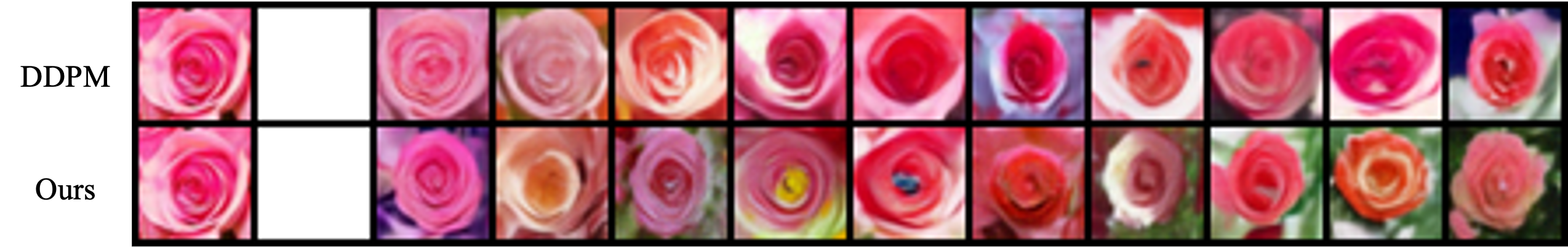}\\[1ex]

    \includegraphics[trim=0 0 0 0, clip, width=1.0\textwidth]{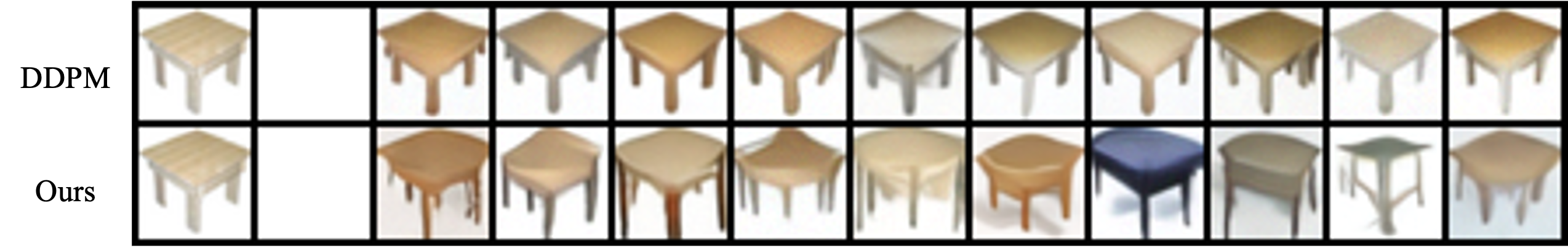}\\[1ex]

    \includegraphics[trim=0 0 0 0, clip, width=1.0\textwidth]{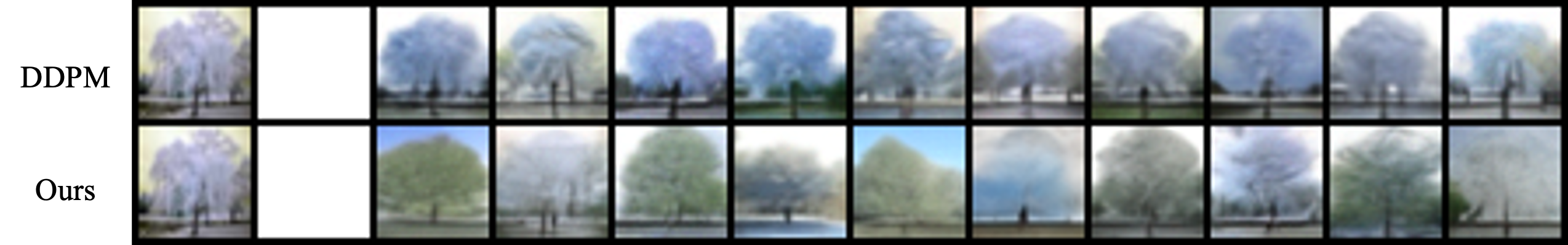}\\[1ex]

    \caption{
    To see overfitting on tail classes, we find Top-10 nearest neighbors (3rd to 12th columns sorted by distances) among 1000 synthetic images to an anchor image (Leftmost column) in the training set. KNN is based on $L_2$ distances of Inception V3 embeddings. For each example, the top row is the results of DDPM, and the bottom row shows ours. The nearest neighbors from DDPM show the mode collapse to the training image, while our method generated more diverse images.
    }
    \label{fig:nearest-neighbor}
\end{figure*}

 % Fig.~\ref{fig:cifar100lt} shows synthetic images of baseline methods and our method for CIFAR100-LT tail classes `rose' and `table'.
 % All methods start the denoising process from the same Gaussian noise at corresponding grid cells.
 % Red dashed ellipses highlight the mode collapse issues observed in DDPM.
 % With the same initial noise, our method gives synthetic images with higher diversity and fidelity compared to DDPM.
 % For example, DDPM always generates rose images containing only one rose. %for the rose image in the sixth column of the last row, %CBDM fails to generate a meaningful image while %
 % %
 % Our method can generate an image containing two roses (see the sixth column of the last row).
 % To further clarify, in the `table' class, DDPM repeatedly produces near-identical images as highlighted in red ellipses. In contrast, our method produces a more varied set of table images. This suggests that our method can better capture greater diversity in image generation when compared to DDPM~\cite{ddpm}.
 % %For instance, for the table image shown in the fourth column of the third row, CBDM and OCLT generate a common table with four legs, while our method generates a less common but valid table with only one leg.
 % %
 To further illustrate how the proposed method mitigates the issue of overfitting, we visualize the top-10 nearest neighbors among 1000 synthetic images to an anchor image in the training set for the original DDPM and our method.
 As shown in Fig.~\ref{fig:nearest-neighbor}, DDPM shows the mode collapse to the training image while our method's generated images show much better diversity.
 For example, in the first two rows, DDPM generates repeated vertical worms while our method generates worms with diverse directions. 
 In the 9th-10th rows, DDPM generates tables with repeated modes while our method generates tables with different styles and colors.

% \begin{figure*}[tb!]
% \centering
% \begin{minipage}{1.0\columnwidth}
% \subfloat[Synthetic images from DDPM~\cite{ddpm}. Red ellipses show similar images generated.]{
% \includegraphics[width=0.982\linewidth]{figures/ddpm.png} 
% }
% \end{minipage}

% \begin{minipage}{1.0\columnwidth}
% \subfloat[Synthetic images from CCUA (ours) have more diversity and are less repetitive.]{
% \includegraphics[width=0.982\linewidth]{figures/ours.png} 
% }
% \end{minipage}

% % \subfigure[Synthetic images from CBDM~\cite{cbdm}.]{
% % \includegraphics[width=\linewidth]{figures/cbdm.png} 
% % }
% % \subfigure[Synthetic images from OCLT~\cite{oclt}.]{
% % \includegraphics[width=\linewidth]{figures/oclt.png}
% % }

% % \DeclareGraphicsExtensions.
% \caption{The synthesized results for CIFAR100-LT tail classes ‘rose’ and ‘table’ are shown for our method and baseline methods. All methods initiate reverse processing from the same Gaussian noise for images at corresponding grid cells. Red dashed ellipses highlight mode collapse issues observed in DDPM.
% Overall, our method demonstrates higher diversity and fidelity compared to DDPM baseline. %For instance, in the rose image located in the sixth column of the last row, CBDM fails to generate a meaningful image, while DDPM and OCLT produce images with only one rose. In contrast, our method generates an image containing two roses, clearly exhibiting greater diversity and improved synthesis quality.
%  }
% \label{fig:cifar100lt}
% \end{figure*}

\subsubsection{More Visualization of Reverse Process of DDPM for Different Classes}
\label{sec:reverseprocessing}
\begin{figure*}[t]
    \centering

    % Row 1
    \includegraphics[trim=0 0 1452 0, clip, width=1.0\textwidth]{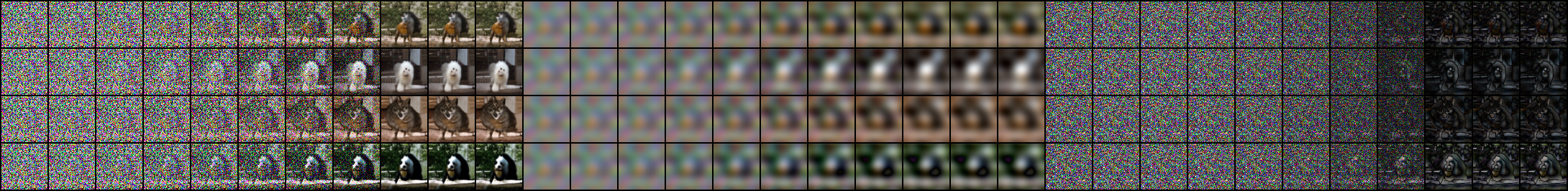}\\[1ex]

    \includegraphics[trim=726 0 726 0, clip, width=1.0\textwidth]{figures/rebuttal_b.png}\\[1ex]

    \includegraphics[trim=1452 0 0 0, clip, width=1.0\textwidth]{figures/rebuttal_b.png}\\[1ex]

    \caption{
    \textbf{Top:} reverse process starting from the same initial Gaussian noise but with different class conditions (2nd-4th rows) or without condition (1st row).
    \textbf{Middle:} low-frequency components of each noisy image.
    \textbf{Bottom:} high-frequency components of each noisy image.
    }
    \label{fig:reverse-2}
\end{figure*}

 To illustrate our observation in Section~\ref{sec:al} more clearly, we decompose $x_t$ into a combination of low-frequency images and high-frequency images, as shown in Fig~\ref{fig:reverse-2}.
 The low-frequency images with the same initial noise are very similar for different classes for the initial steps, which is also observed in prior work~\cite{si2024freeu}.

% \begin{figure*}[t]
%     \centering

%     % Row 1
%     \includegraphics[trim=0 0 1452 0, clip, width=1.0\textwidth]{figures_backup/rebuttal_c.png}\\[1ex]

%     \includegraphics[trim=726 0 726 0, clip, width=1.0\textwidth]{figures_backup/rebuttal_c.png}\\[1ex]

%     \includegraphics[trim=1452 0 0 0, clip, width=1.0\textwidth]{figures_backup/rebuttal_c.png}\\[1ex]

%     \caption{
%     \textbf{Top:} reverse process starting from the same initial Gaussian noise but with different class conditions (2nd-4th rows) or without condition (1st row).
%     \textbf{Middle:} low-frequency components of each noisy image.
%     \textbf{Bottom:} high-frequency components of each noisy image.
%     }
%     \label{fig:reverse-3}
% \end{figure*}

% \begin{figure*}[t]
%     \centering

%     % Row 1
%     \includegraphics[trim=0 0 1452 0, clip, width=1.0\textwidth]{figures_backup/rebuttal_d.png}\\[1ex]

%     \includegraphics[trim=726 0 726 0, clip, width=1.0\textwidth]{figures_backup/rebuttal_d.png}\\[1ex]

%     \includegraphics[trim=1452 0 0 0, clip, width=1.0\textwidth]{figures_backup/rebuttal_d.png}\\[1ex]

%     \caption{
%     \textbf{Top:} reverse process starting from the same initial Gaussian noise but with different class conditions (2nd-4th rows) or without condition (1st row).
%     \textbf{Middle:} low-frequency components of each noisy image.
%     \textbf{Bottom:} high-frequency components of each noisy image.
%     }
%     \label{fig:reverse-4}
% \end{figure*}

\clearpage

\end{document}